\documentclass[10pt,jou rnal,compsoc]{IEEEtran}
\ifCLASSOPTIONcompsoc
  \usepackage[nocompress]{cite}
\else
  \usepackage{cite}
\fi
\ifCLASSINFOpdf
   \usepackage[pdftex]{graphicx}
\else
   \usepackage[dvips]{graphicx}
\fi
\usepackage{amsmath}
\usepackage{algorithm}
\usepackage[noend]{algpseudocode}
\ifCLASSOPTIONcompsoc
 \usepackage[caption=false,font=footnotesize,labelfont=sf,textfont=sf]{subfig}
\else
 \usepackage[caption=false,font=footnotesize]{subfig}
\fi
\usepackage{array}
\usepackage{enumitem}
\usepackage{placeins}
\usepackage{stfloats}
\usepackage{afterpage}
\usepackage{amsfonts}
\usepackage{amssymb}
\usepackage{threeparttable}
\usepackage{tabularx}
\usepackage[colorlinks=true,urlcolor=red]{hyperref}
\makeatletter
\newcommand{\multiline}[1]{%
  \begin{tabularx}{\dimexpr\linewidth-\ALG@thistlm}[t]{@{}X@{}}
    #1
  \end{tabularx}
}
\makeatother
\usepackage[dvipsnames]{xcolor}
\usepackage{booktabs}
\usepackage{multirow}

\title{Unsupervised Multiple-Object Tracking\\ with a Dynamical Variational Autoencoder}
\author{Xiaoyu~Lin,$^1$ Laurent Girin,$^2$ Xavier Alameda-Pineda,$^1$~\IEEEmembership{Senior Member,~IEEE}
\thanks{$^1$ Inria Grenoble Rh\^one-Alpes, Univ. Grenoble-Alpes, France}
\thanks{$^2$ Univ. Grenoble Alpes, Grenoble-INP, GIPSA-lab, France}
\thanks{This research was supported by ANR-3IA MIAI (ANR-19-P3IA-0003), ANR-JCJC ML3RI (ANR-19-CE33-0008-01), H2020 SPRING (funded by EC under GA \#871245).}}

\begin{document}

\IEEEtitleabstractindextext{%
\begin{abstract}
In this paper, we present an unsupervised probabilistic model and associated estimation algorithm for multi-object tracking (MOT) based on a dynamical variational autoencoder (DVAE), called DVAE-UMOT. The DVAE is a latent-variable deep generative model that can be seen as an extension of the variational autoencoder for the modeling of temporal sequences. It is included in DVAE-UMOT to model the objects' dynamics, after being pre-trained on an unlabeled synthetic dataset of single-object trajectories. Then the distributions and parameters of DVAE-UMOT are estimated on each multi-object sequence to track using the principles of variational inference: Definition of an approximate posterior distribution of the latent variables and maximization of the corresponding evidence lower bound of the data likehood function. DVAE-UMOT is shown experimentally to compete well with and even surpass the performance of two state-of-the-art probabilistic MOT models. Code and data are \href{https://gitlab.inria.fr/xilin/dvae-umot-release}{publicly available}.\end{abstract}

\begin{IEEEkeywords}
Unsupervised Learning, Dynamical Variational Autoencoder, Multiple Object Tracking, Probabilistic Tracking \end{IEEEkeywords}}

\maketitle

\IEEEdisplaynontitleabstractindextext

%
\IEEEpeerreviewmaketitle

\IEEEraisesectionheading{\section{Introduction}\label{sec:introduction}}
\IEEEPARstart{M}{ulti-object} tracking (MOT), or multi-target tracking, is a fundamental and very general pattern recognition task. Given an input time-series, the aim of MOT is to recover the trajectories of an unknown number of sources, that might appear and disappear at any point in time~\cite{vo2015multitarget,LUO2021103448}. There are four main challenges associated to MOT, namely: (i) extracting source observations (also called detections) at every time frame, (ii) modeling the dynamics of the sources’ movements, (iii) associating observations to sources consistently over time, and (iv) accounting for birth and death of source trajectories.

In computer vision, the tracking-by-detection paradigm has become extremely popular in the recent years~\cite{4587583,8296962,CIAPARRONE202061,LUO2021103448}. In this context, more and more powerful detection algorithms brought a significant increase of performance \cite{NIPS2015_14bfa6bb,7780460,He_2017_ICCV}. Including motion information also improves the tracking performance~\cite{8451140,Bergmann_2019_ICCV,saleh2021probabilistic}. Among the tracking approaches based on motion models, the linear dynamical model (i.e., constant object velocity) is the most commonly used ~\cite{Bewley_2016,8296962,Bergmann_2019_ICCV}. Further, it makes sense to combine the detection and tracking models together~\cite{Bergmann_2019_ICCV,10.1007/978-3-030-58548-8_28,zhang2021fairmot}.  

While a linear dynamical model is suitable in case of high sampling rate and reasonable object velocity, challenging tracking scenarios with low sampling rate, moving camera, high object velocity, rapid changes of direction, and occlusions are still difficult to tackle with such simple models, thus leading to numerous tracking errors such as false negatives and identity switches. Therefore, more sophisticated nonlinear models using recurrent neural networks (RNNs) to capture the motion information have been proposed~\cite{8237303,8451140,10.5555/3298023.3298181,8651492,8575325,6247892,8451739,6751397}. 

The powerful sequential modeling capabilities of RNNs can help to design more accurate motion models, provided we have enough annotated data to train them. Since obtaining annotations in the MOT framework is a very tedious and resource-consuming task, it is worth investigating the training of complex motion/MOT models in an unsupervised manner, i.e., without human-annotated data. 


In the present paper, we propose a probabilistic \textit{unsupervised} MOT model. We focus on challenges (ii) and (iii) above.
First, we propose to use a dynamical variational autoencoder (DVAE) to model the complex motion patterns of each tracked object. DVAEs are a family of powerful deep probabilistic generative models with latent variables designed for modeling correlated sequences of (multidimensional) data \cite{MAL-089}. They can be seen as a combination of the variational autoencoder (VAE)~\cite{Kingma2014,pmlr-v32-rezende14} with recurrent neural networks (RNNs). The DVAE is pre-trained on a  synthetic dataset of single-object trajectories generated without human annotations. This pre-training is done using the variational inference (VI) methodology, which is classical in the VAE context and consists of defining an approximate posterior distribution of the latent variables and maximizing a lower bound of the data likelihood function. 
Second, the DVAE is combined with an observation-to-object assignment latent variable to solve the MOT task -- hence the name DVAE-UMOT. 
The distributions and parameters of the complete DVAE-UMOT model are estimated on each multi-object sequence to track, using again the VI methodology, which includes here the observation-to-object assignment and the estimation of objects position. Experiments conducted on a dataset derived from MOT17 (which is part of the MOTChallenge~\cite{dendorfer2020motchallenge}) show that our model exhibits a very competitive tracking performance, even for long tracks of 300 frames, and can outperform two recent state-of-the-art MOT models (including a model that was trained using MOT17 annotations) on several MOT metrics. 

The rest of this paper is organised as follows. We discuss the related work in Section~II. The theoretical background of VI and DVAEs is introduced in Section~III. The main developments of the proposed DVAE-UMOT model and algorithm are presented in Section~IV and implementation details are provided in Section~V. Experimental results and comparison with state-of-the-art approaches are presented and discussed in Section~VI. We draw conclusions and discuss the future work in Section~VII.

\section{Related Work}
\par \textbf{Motion models in MOT.} 
In tracking-by-detection MOT algorithms, data association is usually done by extracting features from the video sequences, using them to compute a feature affinity matrix and finally use this matrix to assign detections to targets~\cite{CIAPARRONE202061,LUO2021103448}. A number of appearance models have enabled good tracking performance \cite{8237780,8296962,7410890,7780509,LealTaix2016LearningBT,8099886}. However, models based on visual features only are less robust to similar appearances, occlusions and noisy detections. To solve this problem, several approaches combined motion features with appearance features  to construct more robust MOT models. A Kalman filter was used in \cite{8296962,Bewley_2016,Bergmann_2019_ICCV,zhang2021fairmot} to exploit the motion information, assuming that the targets follow a linear dynamical model, i.e., they have a constant velocity, at least locally (on a portion of the trajectory). In \cite{8907431}, a Kalman filter was integrated into a VI framework so as to combine both audio-visual observations and a motion model together. Nevertheless, linear dynamical models only consider the temporal dependencies between adjacent frames and thus cannot deal efficiently with long-term occlusions. 

With the spreading use of neural networks, more sophisticated nonlinear motion models have been developed. RNNs were used in \cite{10.5555/3298023.3298181} to mimic the classical Bayesian filtering state estimation. The authors of  \cite{8237303} proposed to combine multiple cues such as appearance, motion and interaction cues within an RNN architecture. In \cite{8451140}, RNNs were used to predict the targets motion and tracklet-stitching was used to handle occlusions. All these methods used RNNs for modeling nonlinear temporal dependencies. Still, the capabilities of deterministic RNNs remain limited for long sequences. A recurrent autoregressive network was used in \cite{8354161} to extract long-term temporal dependencies and to learn a probabilistic generative model for multiple object trajectories. The authors of \cite{saleh2021probabilistic} proposed an RNN-based probabilistic autoregressive motion model which can be used to both score tracklet proposals and inpaint tracklets in case of occlusion. These works adopted a probabilistic point of view and constructed more flexible motion models. However, these motion models have not been combined with the MOT algorithm in a principled way. In our proposed model, we adopt a complete probabilistic perspective. We propose to use the DVAE models to capture the long-term temporal dependencies of object positions and combine them with an observation-to-object assignment variable within a VI framework. In addition to estimate the assignment probabilities, the proposed DVAE-UMOT model can also filter out detection noise to form smooth trajectories. The powerful sequential modeling capability of DVAEs guarantees the accuracy of long-term tracking.
\par \textbf{Unsupervised MOT.} Compared with the relatively large number of supervised MOT methods, unsupervised MOT methods are much less explored in the literature and differ a lot from one to another. The authors of \cite{Luiten_2020_WACV} firstly grouped objects into short tracklets using their spatio-temporal consistency. And then long-term consistent tracks were formed based on the objects' visual similarity, using a tracklet-based Forest Path Cutting data association algorithm. The authors of \cite{karthik2020simple} followed a similar line, using a re-identification model trained with noisy-labeled data to extract the appearance features used to form the long-term tracks. Another branch of unsupervised models is based on geometric rendering and visual reconstruction. The authors of \cite{8953199} proposed a tracking-by-animation framework, which first tracks objects from video frames and then renders tracker outputs into reconstructed frames. 
A VAE-based spatially-invariant label-free object tracking model was proposed in \cite{Crawford_Pineau_2020}. In their work, the VAE encoder, which takes as input the sequential images, consists of three modules: a discovery module to detect newly birthed objects, a propagation module to update the attributes of existing objects, and a selection module to select objects and deal with track death. The VAE decoder is a rendering module that reconstructs the video sequence. The model is trained in an unsupervised way without using manual annotations. In our proposed model, we focus on the dynamical modeling of objects in MOT. And we only exploit the motion information without using appearance features. In this context, the term \textit{unsupervised} means that we do not use any manually annotated ground-truth object position sequence to train the model.
\par \textbf{DVAE models.} Firstly proposed in \cite{Kingma2014,pmlr-v32-rezende14}, the VAE is a powerful deep probabilistic generative model which learns a low-dimensional latent space underlying the generation of high-dimensional complex data. It has been successfully applied in many different fields \cite{10.5555/3157096.3157360,10.1145/3097983.3098077,pmlr-v48-maaloe16}. The original VAE model does not include sequential modeling. The DVAE models integrated RNNs into the VAE structure to build deep probabilistic generative models for sequential data \cite{MAL-089}. Similar to the VAE, DVAE models are also based on an encoder/decoder structure, here with a sequence of latent variables associated to a sequence of observed data. As more largely discussed in Section~\ref{subsec:dvae}, different temporal dependencies between the observed and latent variables result in different DVAE models. 
DVAEs have been applied for different tasks, e.g., the disentanglement of speech factors of variation \cite{pmlr-v80-yingzhen18a} and speech enhancement in noise \cite{9053164}. We show in this paper that the DVAE models can also be successfully used for MOT. To our knowledge, this is the first time DVAEs are used for this task. 

\section{Methodological Background}
\label{sec:Background}

\subsection{Latent variable generative models and variational inference}
\label{subsec:LVGM-VI}

The methodology of the proposed MOT algorithm lies under the very broad umbrella of latent variable generative models (LVGMs) and associated variational inference (VI) methodology~\cite{10.1561/2200000001,10.5555/1162264}. 
An LVGM depicts the relationship between an observed $\mathbf{o}$ and a latent $\mathbf{h}$ random (vector) variable from which $\mathbf{o}$ is assumed to be generated. We consider an LVGM defined via a \textit{parametric} joint probability distribution $p_\theta(\mathbf{o},\mathbf{h})$, where $\theta$ denotes the set of parameters. In a general manner, we are interested in two problems closely related to each other: 1) estimate the parameters $\theta$ that maximize the observed data (marginal) likelihood $p_\theta(\mathbf{o})$, and 2) derive the posterior distribution $p_\theta(\mathbf{h}|\mathbf{o})$ so as to infer the latent variable $\mathbf{h}$ from the observation $\mathbf{o}$.

A prominent tool to estimate $\theta$ is the family of expectation-maximisation (EM) algorithms, that maximizes the following lower bound of the marginal likelihood, called the evidence lower-bound (ELBO):
\begin{equation}
    \mathcal{L}(\theta,q;\mathbf{o}) = \mathbb{E}_{q(\mathbf{h}|\mathbf{o})}\bigg[\log\frac{p_\theta(\mathbf{o},\mathbf{h})}{q(\mathbf{h}|\mathbf{o})}\bigg] \leq \log p_\theta(\mathbf{o}), \label{eq:log-marginal-1}
\end{equation}
where $q(\mathbf{h}|\mathbf{o})$ is a distribution on $\mathbf{h}$ conditioned on $\mathbf{o}$.
The EM algorithm is an iterative alternate  optimisation procedure that maximizes the ELBO w.r.t.\ the distribution $q$ (E-step) and the parameters $\theta$ (M-step). Maximizing the ELBO  w.r.t.\ $q$ is equivalent to minimising the Kullback-Leibler divergence (KLD) between $q(\mathbf{h}|\mathbf{o})$ and the exact posterior distribution $p_\theta(\mathbf{h}|\mathbf{o})$ \cite{10.5555/1162264}. When  $p_{\theta}(\mathbf{h}|\mathbf{o})$ is computationally tractable, it is optimal to choose $q(\mathbf{h}|\mathbf{o})=p_{\theta}(\mathbf{h}|\mathbf{o})$, then the ELBO is tight and the EM is called ``exact.'' Otherwise, the optimization w.r.t.\ $q$ is constrained within a given family of computationally tractable distributions and the bound is not tight anymore. We then have to step in the VI framework.

A first family of VI approaches is the structured mean-field method~\cite{parisi1988statistical} which consists in splitting $\mathbf{h}$ into a set of disjoint variables $\mathbf{h}=(\mathbf{h}_1, ..., \mathbf{h}_M)$. The approximate posterior distribution $q$ is thus assumed to factorise over this set, i.e., $q(\mathbf{h}|\mathbf{o})=\textstyle \prod_{i=1}^M q_i(\mathbf{h}_i|\mathbf{o})$, which leads to the following optimal factor, given the parameters computed at the previous M-step, $\theta^{\textrm{old}}$:
\begin{equation}
    q_i^*(\mathbf{h}_i|\mathbf{o}) \propto \exp \Big(\mathbb{E}_{\prod_{j\neq i}q_{j}(\mathbf{h}_j|\mathbf{o})}\big[\log p_{\theta^{\textrm{old}}}(\mathbf{o},\mathbf{h})\big] \Big).
    \label{eq:optimal-factorized-posterior-general}
\end{equation}
Since each factor is expressed as a function of the others, this formula is iteratively applied in the E-step of the EM algorithm until some convergence criterion is met. 

A second approach is to rely on amortised inference \cite{JMLR:v14:hoffman13a}, where a set of shared parameters is used to compute the parameters of the approximate posterior distribution. A very well-known example is the variational autoencoder (VAE) \cite{Kingma2014, pmlr-v32-rezende14}. For a reason that will become clear in Section~\ref{sec:U-MOT}, let us here denote by $\mathbf{s}$ the observed variable and by $\mathbf{z}$ the latent one. In a VAE, the joint distribution $ p_{\theta}(\mathbf{s}, \mathbf{z}) = p_{\theta}(\mathbf{s}|\mathbf{z})p(\mathbf{z})$ is defined via the prior distribution on $\mathbf{z}$, generally chosen as the standard Gaussian distribution $p(\mathbf{z}) = \mathcal{N}(\mathbf{z};\mathbf{0}, \mathbf{I})$, and via the conditional distribution on $\mathbf{s}$, generally chosen as a Gaussian with diagonal covariance matrix    $p_{\theta}(\mathbf{s}|\mathbf{z}) = \mathcal{N}\big(\mathbf{s}; \boldsymbol{\mu}_{\theta}(\mathbf{z}), \textrm{diag}(\boldsymbol{v}_{\theta}(\mathbf{z}))\big)$. The mean and variance vectors $\boldsymbol{\mu}_{\theta}(\mathbf{z})$ and $\boldsymbol{v}_{\theta}(\mathbf{z})$ are nonlinear functions of $\mathbf{z}$ provided by a 
deep neural network (DNN), called the \textit{decoder network}, taking $\mathbf{z}$ as input. $\theta$ is here the (amortized) set of parameters  of the DNN.
The posterior distribution $p_{\theta}(\mathbf{z}| \mathbf{s})$ corresponding to this model does not have an analytical expression and it is approximated by a Gaussian distribution with diagonal covariance matrix $q_{\phi}(\mathbf{z}|\mathbf{s}) = \mathcal{N}\big(\mathbf{z}; \boldsymbol{\mu}_{\phi}(\mathbf{s}), \textrm{diag}(\boldsymbol{v}_{\phi}(\mathbf{s}))\big)$,
where the mean and variance vectors $\boldsymbol{\mu}_{\phi}(\mathbf{s})$ and $\boldsymbol{v}_{\phi}(\mathbf{s})$ are non-linear functions of $\mathbf{s}$ implemented by another DNN called the \textit{encoder network} and parameterized by $\phi$. The ELBO is here given by:
\begin{align}
    \mathcal{L}(\theta, \phi; \mathbf{s}) 
    & = \mathbb{E}_{q_{\phi}(\mathbf{z}|\mathbf{s})}\big[\log p_{\theta}(\mathbf{s}, \mathbf{z}) - \log q_{\phi}(\mathbf{z}|\mathbf{s})\big].\label{eq:VAE-ELBO-1}
\end{align}
In practice, the ELBO is jointly optimized w.r.t. $\theta$ and $\phi$ using a combination of stochastic gradient descent (SGD) and sampling \cite{Kingma2014}. This is in contrast with the EM algorithm where $q$ and $\theta$ are  optimized alternatively.

\subsection{From VAE to DVAE} \label{subsec:dvae}
\par In the VAE, each observed data vector $\mathbf{s}$ is considered independently of the other data vectors. The dynamical variational autoencoders (DVAEs) are a class of models that extend and generalize the VAE to model sequences of data vectors correlated in time \cite{MAL-089}. Roughly speaking, DVAE models combine a VAE with temporal models such as recurrent neural networks (RNNs) and/or state space models (SSMs). They can be used to model sequences of object bounding boxes/positions and are therefore suitable for MOT. 
\par Let $\mathbf{s}_{1:T}=\{\mathbf{s}_t\}_{t=1}^T$ and $\mathbf{z}_{1:T}=\{\mathbf{z}_t\}_{t=1}^T$ be a discrete-time sequence of observed and latent vectors, respectively, and let $\mathbf{p}_t=\{\mathbf{s}_{1:t-1},\mathbf{z}_{1:t-1}\}$ denote the set of past observed and latent vectors at time $t$. Using the chain rule, the most general DVAE generative distribution can be written as the following causal generative process:
\begin{equation}
    p_{\theta}(\mathbf{s}_{1:T}, \mathbf{z}_{1:T}) = \prod_{t=1}^T p_{\theta_{\mathbf{s}}}(\mathbf{s}_t|\mathbf{p}_{t}, \mathbf{z}_{t})p_{\theta_{\mathbf{z}}}(\mathbf{z}_t|\mathbf{p}_{t}),
    \label{eq:DVAE-joint-chained}
\end{equation}
where $p_{\theta_{\mathbf{s}}}(\mathbf{s}_t|\mathbf{p}_{t}, \mathbf{z}_{t})$ and $p_{\theta_{\mathbf{z}}}(\mathbf{z}_t|\mathbf{p}_{t})$ are arbitrary generative distributions, which parameters are provided sequentially by RNNs taking the respective conditioning variables as inputs. A common choice is to use Gaussian distributions with diagonal covariance matrices: 
\begin{equation}
    p_{\theta_{\mathbf{s}}}(\mathbf{s}_{t} | \mathbf{p}_{t}, \mathbf{z}_{t}) = \mathcal{N}\big(\mathbf{s}_{t}; \boldsymbol{\mu}_{\theta_{\mathbf{s}}}(\mathbf{p}_{t}, \mathbf{z}_{t}), \textrm{diag}(\boldsymbol{v}_{\theta_{\mathbf{s}}}(\mathbf{p}_{t}, \mathbf{z}_{t}))\big)
    \label{eq:DVAE-generative-distr-s}
\end{equation}
\begin{equation}
    p_{\theta_{\mathbf{z}}}(\mathbf{z}_{t} | \mathbf{p}_{t}) = \mathcal{N}\big(\mathbf{z}_{t}; \boldsymbol{\mu}_{\theta_{\mathbf{z}}}(\mathbf{p}_{t}), \textrm{diag}(\boldsymbol{v}_{\theta_{\mathbf{z}}}(\mathbf{p}_{t}))\big).
    \label{eq:DVAE-generative-distr-z}
\end{equation}
It can be noted that the distribution of $\mathbf{z}_t$ is more complex than the standard Gaussian used in the vanilla VAE. Also, the different models belonging to the DVAE class differ in the possible conditional independence assumptions that can be made in (\ref{eq:DVAE-joint-chained}).

As for the VAE, the exact posterior distribution $p_{\theta}(\mathbf{z}_{1:T}| \mathbf{s}_{1:T})$ corresponding to the DVAE generative model is not analytically tractable. Again, an inference model $q_{\phi_{\mathbf{z}}}(\mathbf{z}_{1:T}| \mathbf{s}_{1:T})$ is defined to approximate the exact posterior distribution and factorises as:
\begin{equation}
    q_{\phi_{\mathbf{z}}}(\mathbf{z}_{1:T}| \mathbf{s}_{1:T}) = \prod_{t=1}^Tq_{\phi_{\mathbf{z}}}(\mathbf{z}_t| \mathbf{q}_{t}),
    \label{eq:DVAE-inference-model-chained}
\end{equation}
where $\mathbf{q}_t=\{\mathbf{z}_{1:t-1},\mathbf{s}_{1:T}\}$ denotes the set of past latent variables and all observations. Again, the Gaussian distribution with diagonal covariance matrix is generally used:
\begin{equation}
    q_{\phi_{\mathbf{z}}}(\mathbf{z}_{t} | \mathbf{q}_{t} ) = \mathcal{N}\big(\mathbf{z}_{t}; \boldsymbol{\mu}_{\phi_{\mathbf{z}}}(\mathbf{q}_{t}), \textrm{diag}(\boldsymbol{v}_{\phi_{\mathbf{z}}}(\mathbf{q}_{t}))\big),
    \label{eq:DVAE-inference-model-Gaussian}
\end{equation}
where the mean and variance vectors are provided by an RNN (the encoder network) taking $\mathbf{q}_{t}$ as input and parameterized by $\phi$. With the most general generative model defined in \eqref{eq:DVAE-joint-chained}, the conditional distribution in \eqref{eq:DVAE-inference-model-chained} cannot be simplified. However, if conditional independence assumptions have been made in  (\ref{eq:DVAE-joint-chained}), the dependencies in $q_{\phi_{\mathbf{z}}}(\mathbf{z}_t|\mathbf{z}_{1:t-1}, \mathbf{s}_{1:T})$ might be simplified using the D-separation method \cite{10.5555/1162264,geiger1990identifying}, see \cite{MAL-089} for details. In addition, we can force the inference model to be causal by replacing $\mathbf{s}_{1:T}$ with $\mathbf{s}_{1:t}$. This is particularly suitable for on-line processing. In the rest of the paper, we will use the causal inference model, i.e., $\mathbf{q}_t=\{\mathbf{z}_{1:t-1},\mathbf{s}_{1:t}\}$ in \eqref{eq:DVAE-inference-model-chained} and \eqref{eq:DVAE-inference-model-Gaussian}.

\par Similar to the VAE, and following the general VI principle, the DVAE model is also trained by maximizing the ELBO with a combination of SGD and sequential sampling. The ELBO has here the following general form \cite{MAL-089}:
\begin{align}
    \mathcal{L}(\theta_{\mathbf{s}}, \theta_{\mathbf{z}}, \phi_{\mathbf{z}}; \mathbf{s}_{1:T}) 
    & = \mathbb{E}_{q_{\phi_{\mathbf{z}}}(\mathbf{z}_{1:T}|\mathbf{s}_{1:T})}\big[\log p_{\theta}(\mathbf{s}_{1:T}, \mathbf{z}_{1:T})\nonumber\\
    & \qquad - \log q_{\phi_{\mathbf{z}}}(\mathbf{z}_{1:T}|\mathbf{s}_{1:T})\big] \label{eq:DVAE-ELBO-general-form-a}.
\end{align}


\section{DVAE-UMOT Model and Algorithm}
\label{sec:U-MOT}


\subsection{Problem formulation and notations}
\label{sec:problem}

Let us consider a video sequence containing $N$ moving objects that we want to track over time. Let $n\in \{1, ..., N\}$ denote the object index and let $\mathbf{s}_{tn}$ be here the true (unknown) \textit{position vector} of object $n$ at time frame $t$. As commonly done in the computer vision community, we use a bounding box to characterize an object position, and $\mathbf{s}_{tn} = (s^\textsc{l}_{tn}, s^\textsc{t}_{tn}, s^\textsc{r}_{tn}, s^\textsc{b}_{tn}) \in \mathbb{R}^4$ is the coordinates vector of the top-left and bottom-right points of the bounding box that represents the true position of object $n$. As our model follows the tracking-by-detection paradigm, the observations that we use here are the detected bounding boxes provided by a customized detector such as FRCNN~\cite{NIPS2015_14bfa6bb}. Let us denote by $K_t$ the total number of detected bounding boxes at time frame $t$, which can vary over time. We denote by $\mathbf{o}_{tk} = (o^\textsc{l}_{tk}, o^\textsc{t}_{tk}, o^\textsc{r}_{tk}, o^\textsc{b}_{tk}) \in \mathbb{R}^4$, $k\in \{1, ...,K_t\}$, the coordinates vector of the top-left and bottom-right points of the $k$-th detected bounding box at frame $t$. 

The MOT problem consists in estimating the position vector sequence 
$\mathbf{s}_{1:T,n}=\{\mathbf{s}_{tn}\}_{t=1}^T$, for each object $n$, from the complete set of observations $\mathbf{o}_{1:T,1:K_t}=\{\mathbf{o}_{tk}\}_{t=1,k=1}^{T,K_t}$. To solve this problem, we define two additional sets of latent variables. First, for each object $n$ at time frame $t$, we define a latent variable $\mathbf{z}_{tn} \in \mathbb{R}^L$, which is associated to $\mathbf{s}_{tn}$ through a DVAE model. This DVAE model, which is identical for all objects, is used to model the dynamics of each object in the analyzed video and plugged into the proposed MOT model. In a VAE or DVAE, the latent dimension $L$ is usually smaller than the dimension of the observed vector, in order to obtain a compact data representation. In the present work, the input dimension, which equals 4, is already small, so that we set the same dimension for $\mathbf{z}_{tn}$; that is, $L = 4$. Second, for each observation $\mathbf{o}_{tk}$, we define a discrete observation-to-object assignment variable $w_{tk}$ taking its value in $\{1, ..., N\}$. $w_{tk}=n$ means that observation $k$ at time frame $t$ is assigned to object $n$. This variable is used to form the tracklets. Each tracklet is a consistent object trajectory, i.e., a sequence of estimated position vectors consistent over time and assigned to the same object.

Hereinafter, to simplify the notations, we use ``:'' as a shortcut subscript for the set of all values of the corresponding index. For example, $\mathbf{s}_{:,n}=\mathbf{s}_{1:T,n}$ is the complete trajectory of object $n$ and $\mathbf{s}_{t,:}=\mathbf{s}_{t,1:N}$ is the set of position vectors for all objects at time frame $t$. 
All notations are summarized in Table~\ref{table:notations}. 

\begin{table}[!t]
\centering
\caption{Summary of the Notations}
\label{table:notations}
 \begin{tabular}{p{2.7cm}p{5.4cm}} 
 \toprule
 Variable notation & Definition \\
 \midrule
 $T$, $t\in \{1,\ldots,T\}$ & Sequence length and frame index \\
 $N$, $n\in \{1,\ldots,N\}$ & Total number of objects and object index \\
 $K_t$, $k\in \{1,\ldots,K_t\}$ & Number of observations at $t$, and obs. index \\
 $\mathbf{s}_{tn} \in \mathbb{R}^4$ & True position of object $n$ at time $t$ \\
 $\mathbf{z}_{tn} \in \mathbb{R}^L$ & Latent variable of object $n$ at time $t$ \\
 $\mathbf{o}_{tk} \in \mathbb{R}^4$ & Observation $k$ at time $t$ \\
 $w_{tk} \in \{1,\ldots,N\}$ & Assignment of observation $k$ at time $t$ \\
 $\mathbf{s}_{:, n}=\mathbf{s}_{1:T, n}$ & Position sequence of object $n$ \\
 $\mathbf{s}_{t, :}=\mathbf{s}_{t, 1:N}$ & Position of all objects at time $t$ \\
 $\mathbf{s}=\mathbf{s}_{1:T, 1:N}$ & Set of all object positions \\
 $\mathbf{z}_{:, n}$, $\mathbf{z}_{t, :}$, $\mathbf{z}$ & Analogous for the latent variable \\
 $\mathbf{o}=\mathbf{o}_{1:T, 1:K_t}$ & Set of all observations \\
 $\mathbf{w}=\mathbf{w}_{1:T, 1:K_t}$ & Set of all assignment variables \\
 \bottomrule
 \end{tabular}
\end{table}

\subsection{General principle of the proposed solution}
\label{sec:general-principle}

The general methodology of DVAE-UMOT is to define a parametric joint distribution of all variables $p_{\theta}(\mathbf{o}, \mathbf{s}, \mathbf{z}, \mathbf{w})$, then estimate its parameters $\theta$ and (an approximation of) the corresponding posterior distribution $p_{\theta}(\mathbf{s}, \mathbf{z}, \mathbf{w}|\mathbf{o})$, from which we can deduce an estimate of $\mathbf{s}_{1:T,n}$ for each object $n$. The proposed DVAE-UMOT generative model $p_{\theta}(\mathbf{o}, \mathbf{s}, \mathbf{z}, \mathbf{w})$ is presented in Section~\ref{sec:generative_model}. As briefly stated above, it integrates the DVAE generative model \eqref{eq:DVAE-joint-chained}--\eqref{eq:DVAE-generative-distr-z} to model the objects dynamics and does not use any human-annotated data for training. 

As it is usually the case in (D)VAE-based generative models, both the exact posterior distribution $p_{\theta}(\mathbf{s}, \mathbf{z}, \mathbf{w}|\mathbf{o})$ and the marginalisation of the joint distribution $p_{\theta}(\mathbf{o}, \mathbf{s}, \mathbf{z}, \mathbf{w})$ are analytically intractable. Therefore we cannot directly use an exact EM algorithm and we resort to VI. We 
propose the following strategy, inspired by the structured mean-field method that we summarized in Section~\ref{subsec:LVGM-VI}, with $\mathbf{h}$ being here equal to  $\{\mathbf{s}, \mathbf{z}, \mathbf{w}\}$. 
 In Section~\ref{sec:MOT-inference-model}, we define an approximate posterior distribution $q_{\phi}(\mathbf{s}, \mathbf{z}, \mathbf{w}|\mathbf{o})$ that partially factorizes over $\{\mathbf{s}, \mathbf{z}, \mathbf{w}\}$. As a parallel to the proposed DVAE-UMOT generative model integrating the DVAE generative model, this approximate DVAE-UMOT posterior distribution includes the DVAE inference model. Finally, in Section~\ref{sec:MOT-algorithm}, we present an EM-like iterative algorithm for jointly deriving the terms of the  inference model (other than the DVAE terms) and estimating the parameters of the complete DVAE-UMOT model (i.e., training this model), based on the maximization of the corresponding ELBO. As we will see, this training is done directly on each multi-object sequence to process (i.e., a test MOT sequence) and does not require previous supervised training with a labeled dataset of multi-object training sequences. It only requires to pre-train the DVAE model, which is done on a dataset of synthetic \emph{single-object} sequences. This is possible because the DVAE is a single-object model, which is applied $n$ times in parallel in the proposed MOT algorithm to process an $n$-object scene.

\subsection{Generative model}
\label{sec:generative_model}
\par In this section, we define the proposed DVAE-UMOT generative model; that is, we specify the joint distribution of observed and latent variables $p_{\theta}(\mathbf{o}, \mathbf{w}, \mathbf{s}, \mathbf{z})$. We assume that the observation variable $\mathbf{o}$ only depends on $\mathbf{w}$ and $\mathbf{s}$, while the assignment variable $\mathbf{w}$ is a priori independent of the other variables. Applying the chain rule and these conditional dependency assumptions, the joint distribution can be factorised as follows:
\begin{equation}
    p_{\theta}(\mathbf{o}, \mathbf{w}, \mathbf{s}, \mathbf{z}) = p_{\theta_{\mathbf{o}}}(\mathbf{o}|\mathbf{w}, \mathbf{s})p_{\theta_{\mathbf{w}}}(\mathbf{w})p_{\theta_{\mathbf{s}\mathbf{z}}}(\mathbf{s}, \mathbf{z}).\label{eq:MOT-generative-factorized}
\end{equation}

\noindent \textbf{Observation model:} We assume that the observations are conditionally independent through time and independent to each other, that is to say, at any time frame $t$, the observation  $\mathbf{o}_{tk}$ only depends on its corresponding assignment $w_{tk}$ and position vector at the same time frame. The observation model $p_{\theta_{\mathbf{o}}}(\mathbf{o}|\mathbf{w}, \mathbf{s})$ can thus be factorised as:\footnote{In this equation, we use $\mathbf{s}_{t, :}$ and not $\mathbf{s}_{tn}$, since the value of $w_{tk}$ is not specified.}
\begin{equation}
    p_{\theta_{\mathbf{o}}}(\mathbf{o}|\mathbf{w}, \mathbf{s}) = \prod_{t=1}^T\prod_{k=1}^{K_t} p_{\theta_{\mathbf{o}}}(\mathbf{o}_{tk} | w_{tk}, \mathbf{s}_{t, :}).\label{eq:MOT-obs-factorized}
\end{equation}
Given the value of the assignment variable, the distribution $p(\mathbf{o}_{tk} | w_{tk}, \mathbf{s}_{t, :})$ is modeled by a Gaussian distribution:
\begin{equation}
  p_{\theta_{\mathbf{o}}}(\mathbf{o}_{tk} | w_{tk} = n, \mathbf{s}_{tn}) = \mathcal{N}(\mathbf{o}_{tk}; \mathbf{s}_{tn}, \boldsymbol{\Phi}_{tk}).\label{eq:MOT-obs-Gaussian}
\end{equation}
This choice is motivated by the fact that a detected bounding box is generally not very far from the corresponding true bounding box. In other words, $\mathbf{o}_{tk}$ is a noisy version of $\mathbf{s}_{tn}$ (assuming $w_{tk} = n$). $\boldsymbol{\Phi}_{tk} \in \mathbb{R}^{4\times 4}$ is the covariance matrix of the observation noise. 


\vspace{0.2cm}
\noindent \textbf{Assignment model:}  Similarly, we assume that, a priori, the assignment variables are independent across time and observations:
\begin{equation}
    p_{\theta_{\mathbf{w}}}(\mathbf{w}) = \prod_{t=1}^T\prod_{k=1}^{K_t} p_{\theta_{\mathbf{w}}}(w_{tk}).
    \label{eq:MOT-assign-factorized}
\end{equation}
For each time frame $t$ and each observation $k$, the assignment variable $w_{tk}$ is assumed to follow a uniform prior distribution:
\begin{equation}
    p_{\theta_{\mathbf{w}}}(w_{tk}) = \frac{1}{N}.
    \label{eq:MOT-assign-uniform}
\end{equation}

\vspace{0.2cm}
\noindent \textbf{Dynamical model:} Finally, $p_{\theta_{\mathbf{s}\mathbf{z}}}(\mathbf{s}, \mathbf{z})$ is modeled with a DVAE. The different tracked objects are assumed to be independent of each other. This implies that in the present work we do not consider the interactions among objects during tracking. More complex tracking models including object interaction are beyond the scope of this paper. With this assumption, the joint distribution of all objects position and corresponding latent variable $p_{\theta_{\mathbf{s}\mathbf{z}}}(\mathbf{s}, \mathbf{z})$ can be factorised across objects as:
\begin{equation}
    p_{\theta_{\mathbf{s}\mathbf{z}}}(\mathbf{s}, \mathbf{z}) = \prod_{n=1}^N p_{\theta_{\mathbf{s}\mathbf{z}}}(\mathbf{s}_{:, n}, \mathbf{z}_{:, n}),
    \label{eq:MOT-DVAE-factorized-over-n}
\end{equation}
where $p_{\theta_{\mathbf{s}\mathbf{z}}}(\mathbf{s}_{:, n}, \mathbf{z}_{:, n})$ is the DVAE model defined in \eqref{eq:DVAE-joint-chained}--\eqref{eq:DVAE-generative-distr-z} and applied to  $\mathbf{s}_{:, n}$ and $\mathbf{z}_{:, n}$ (defining $\mathbf{p}_{t,n}=\{\mathbf{s}_{1:t-1, n}, \mathbf{z}_{1:t-1, n}\}$).\footnote{Here we denote the DVAE parameters by $\theta_{\mathbf{s}\mathbf{z}}$ instead of $\theta$ in \eqref{eq:DVAE-joint-chained} to differentiate the DVAE parameters from the other MOT model parameters.} 
In this work, we assume that the dynamics of all objects are modeled with the same DVAE encoder and decoder. 

Overall, the generative model parameters to be estimated are: $\theta = \{\theta_{\mathbf{o}} = \{\boldsymbol{\Phi}_{tk}\}_{t,k=1}^{T, K_t}, \theta_{\mathbf{s}}, \theta_{\mathbf{z}}\}$ (note that $\theta_{\mathbf{w}}=\emptyset$). 


\subsection{Inference model}
\label{sec:MOT-inference-model}

%
The exact posterior distribution corresponding to the generative model described in Section~\ref{sec:generative_model} is neither analytically nor computationally tractable \cite{MAL-089,8907431}. Therefore, we propose the following factorised approximation that leads to a computationally tractable inference model:
\begin{equation}
    q_{\phi}(\mathbf{s}, \mathbf{z}, \mathbf{w} | \mathbf{o}) = q_{\phi_{\mathbf{w}}}(\mathbf{w} | \mathbf{o})q_{\phi_{\mathbf{z}}}(\mathbf{z} | \mathbf{s} )q_{\phi_{\mathbf{s}}}(\mathbf{s} | \mathbf{o}),\label{eq:inf_approximation}
\end{equation}
where $q_{\phi_{\mathbf{z}}}(\mathbf{z} | \mathbf{s} )$ corresponds to the inference model of the DVAE and the optimal distributions $q_{\phi_{\mathbf{s}}}(\mathbf{s} | \mathbf{o})$ and $q_{\phi_{\mathbf{w}}}(\mathbf{w} | \mathbf{o})$ are derived below in the E-step of the DVAE-UMOT algorithm. Note that, the factorization \eqref{eq:inf_approximation} is
inspired by the structured mean-field method \cite{parisi1988statistical}, since we break the posterior dependency between $\mathbf{w}$ and $(\mathbf{s},\mathbf{z})$. However, we keep the dependency between $\mathbf{s}$ and $\mathbf{z}$ at inference time since it is the essence of the DVAE. In addition, we assume that the posterior distribution of the DVAE latent variable is independent for each tracked object, so that we have:
\begin{equation}
    q_{\phi_{\mathbf{z}}}(\mathbf{z} | \mathbf{s}) = \prod_{n=1}^N q_{\phi_{\mathbf{z}}}(\mathbf{z}_{:,n} | \mathbf{s}_{:,n} ), \label{eq:qz_factorises}
\end{equation}
where $q_{\phi_{\mathbf{z}}}(\mathbf{z}_{:,n} | \mathbf{s}_{:,n} )$ is given by    \eqref{eq:DVAE-inference-model-chained} and \eqref{eq:DVAE-inference-model-Gaussian} applied to $\{\mathbf{z}_{:,n},\mathbf{s}_{:,n}\}$.
This is coherent with the generative model, where we assumed that the dynamics of the various tracked objects are independent of each other. 

\subsection{DVAE-UMOT algorithm and solution}
\label{sec:MOT-algorithm}

To derive the posterior distributions $q_{\phi_{\mathbf{s}}}(\mathbf{s}|\mathbf{o})$ and $q_{\phi_{\mathbf{w}}}(\mathbf{w}|\mathbf{o})$ and the parameters of all the other involved distributions, we cannot directly use the generic solution of the structured mean-field method \eqref{eq:optimal-factorized-posterior-general} because \eqref{eq:inf_approximation} does not factorise in disjoint latent  variable sets (in other words, $q_{\phi_{\mathbf{z}}}(\mathbf{z} | \mathbf{s} )$ in \eqref{eq:inf_approximation} is conditioned on $\mathbf{s}$ and not on $\mathbf{o}$). 
We thus have to go back to the fundamentals of VI and iteratively maximize the ELBO of the DVAE-MOT model defined by:
\begin{equation}
    \mathcal{L}(\theta, \phi; \mathbf{o}) = \mathbb{E}_{q_{\phi}(\mathbf{s}, \mathbf{z}, \mathbf{w} | \mathbf{o})}[\log p_{\theta}(\mathbf{o}, \mathbf{s}, \mathbf{z}, \mathbf{w}) - \log q_{\phi}(\mathbf{s}, \mathbf{z}, \mathbf{w} | \mathbf{o})].\label{eq:MOT-ELBO}
\end{equation}
By injecting \eqref{eq:MOT-generative-factorized} and \eqref{eq:inf_approximation} into \eqref{eq:MOT-ELBO}, we can develop $\mathcal{L}(\theta, \phi; \mathbf{o})$ as follows:
\begin{align}
    \mathcal{L}(\theta, \phi; \mathbf{o})
    & = \mathbb{E}_{q_{\phi_{\mathbf{w}}}(\mathbf{w}|\mathbf{o})q_{\phi_{\mathbf{s}}}(\mathbf{s}|\mathbf{o})}\big[\log p_{\theta_{\mathbf{o}}}(\mathbf{o} | \mathbf{w}, \mathbf{s})\big]\nonumber\\
    & + \mathbb{E}_{q_{\phi_{\mathbf{w}}}(\mathbf{w}|\mathbf{o})}\big[\log p_{\theta_{\mathbf{w}}}(\mathbf{w}) - \log q_{\phi_{\mathbf{w}}}(\mathbf{w}|\mathbf{o}) \big] \nonumber \\ 
    & + \mathbb{E}_{q_{\phi_{\mathbf{s}}}(\mathbf{s}|\mathbf{o})}\Big[\mathbb{E}_{q_{\phi_{\mathbf{z}}}(\mathbf{z} | \mathbf{s})}\big[\log p_{\theta_{\mathbf{s}\mathbf{z}}}(\mathbf{s}, \mathbf{z}) - \log q_{\phi_{\mathbf{z}}}(\mathbf{z} | \mathbf{s})\big]\Big]\nonumber\\ 
    & - \mathbb{E}_{q_{\phi_{\mathbf{s}}}(\mathbf{s}|\mathbf{o})} \big[\log q_{\phi_{\mathbf{s}}}(\mathbf{s}|\mathbf{o})\big].
    \label{eq:elbo}
\end{align}
The ELBO maximization is done by alternatively and iteratively maximizing the different terms corresponding to the various posterior and generative distributions. In the EM terminology, this corresponds to the following E and M steps.

\subsubsection{E-S step}
\label{sec:E-S}

We first consider the computation of the optimal posterior distribution $q_{\phi_{\mathbf{s}}}(\mathbf{s}|\mathbf{o})$. To this aim, we first select the terms in (\ref{eq:elbo}) that depend on $\mathbf{s}$ (the other terms being considered as a constant in this part of the optimization):
\begin{align}
 &\mathcal{L}_{\mathbf{s}}(\theta, \phi; \mathbf{o}) = \mathbb{E}_{q_{\phi_{\mathbf{s}}}(\mathbf{s}|\mathbf{o})}\Big[ \mathbb{E}_{q_{\phi_{\mathbf{w}}}(\mathbf{w}|\mathbf{o})}\big[\log p_{\theta_{\mathbf{o}}}(\mathbf{o} | \mathbf{w}, \mathbf{s})\big]\label{eq:VLB-E-S-step-1}\\ 
 &\quad + \mathbb{E}_{q_{\phi_{\mathbf{z}}}(\mathbf{z} | \mathbf{s})}\big[\log p_{\theta_{\mathbf{s}\mathbf{z}}}(\mathbf{s}, \mathbf{z}) - \log q_{\phi_{\mathbf{z}}}(\mathbf{z} | \mathbf{s})\big] - \log q_{\phi_{\mathbf{s}}}(\mathbf{s}|\mathbf{o})\Big].\nonumber
 \end{align}
Let us define: 
\begin{align}\tilde{p}(\mathbf{s}|\mathbf{o})=\mathcal{C}'\exp \Big( \mathbb{E}_{q_{\phi_{\mathbf{w}}}(\mathbf{w}|\mathbf{o})}\big[\log p_{\theta_{\mathbf{o}}}(\mathbf{o} | \mathbf{w}, \mathbf{s})\big]\nonumber\\ 
\qquad + \mathbb{E}_{q_{\phi_{\mathbf{z}}}(\mathbf{z} | \mathbf{s})}\big[\log p_{\theta_{\mathbf{s}\mathbf{z}}}(\mathbf{s}, \mathbf{z}) - \log q_{\phi_{\mathbf{z}}}(\mathbf{z} | \mathbf{s})\big]\Big),
\end{align}
where $\mathcal{C}'>0$ is the appropriate normalisation constant. \eqref{eq:VLB-E-S-step-1} rewrites:
 \begin{align}
 \mathcal{L}_{\mathbf{s}}(\theta, \phi; \mathbf{o}) = -D_{\textsc{kl}}\big(  q_{\phi_{\mathbf{s}}}(\mathbf{s}|\mathbf{o}) \parallel \tilde{p}(\mathbf{s}|\mathbf{o}) \big) + \mathcal{C},
\end{align}
where $D_{\textsc{kl}}(\cdot|\cdot)$ denotes the Kullback-Leibler divergence (KLD).
Therefore, the optimal distribution is the one minimising the above KLD:
\begin{align}
 q_{\phi_{\mathbf{s}}}(\mathbf{s}|\mathbf{o}) & = \tilde{p}(\mathbf{s}|\mathbf{o})\propto \exp \Big(\mathbb{E}_{q_{\phi_{\mathbf{w}}}(\mathbf{w}|\mathbf{o})}\big[\log p_{\theta_{\mathbf{o}}}(\mathbf{o} | \mathbf{w}, \mathbf{s})\big] \nonumber\\
 &+ \mathbb{E}_{q_{\phi_{\mathbf{z}}}(\mathbf{z} | \mathbf{s})}\big[\log p_{\theta_{\mathbf{s}\mathbf{z}}}(\mathbf{s}, \mathbf{z})-\log q_{\phi_{\mathbf{z}}}(\mathbf{z} | \mathbf{s})\big]\Big).
 \label{eq:optimal_qs}
\end{align}
Since for any pair $(t, k)$, the assignment variable $w_{tk}$ follows a discrete posterior distribution, we can denote the corresponding probability values by:
\begin{equation}
    \eta_{tkn} = q_{\phi_{\mathbf{w}}}(w_{tk} = n|\mathbf{o}_{tk}).
    \label{post-distrib-w}
\end{equation}
These values will be computed in the E-W step below.
The expectation with respect to $q_{\phi_{\mathbf{w}}}(\mathbf{w}|\mathbf{o})$ in \eqref{eq:optimal_qs} can be calculated using these values. However, the expectation with respect to $q_{\phi_{\mathbf{z}}}(\mathbf{z}|\mathbf{s})$ cannot be calculated in closed form. As usually done in the (D)VAE methodology, it is thus replaced by a Monte Carlo estimate using sampled sequences drawn from the DVAE inference model at the previous iteration (see Section~\ref{subsec:sampling-order}). Replacing the distributions in \eqref{eq:optimal_qs} with \eqref{eq:MOT-obs-factorized}, \eqref{eq:MOT-DVAE-factorized-over-n}, and \eqref{eq:qz_factorises}, and calculating the expectations with respect to $q_{\phi_{\mathbf{w}}}(\mathbf{w}|\mathbf{o})$ and $q_{\phi_{\mathbf{z}}}(\mathbf{z} | \mathbf{s})$, we find that $q_{\phi_{\mathbf{s}}}(\mathbf{s}|\mathbf{o})$ factorises with respect to $n$ as follows:
\begin{equation}
    q_{\phi_{\mathbf{s}}}(\mathbf{s}|\mathbf{o}) 
    = \prod_{n=1}^N q_{\phi_{\mathbf{s}}}(\mathbf{s}_{:, n}|\mathbf{o}). \label{e37}
\end{equation}

Each of these factors corresponds to the posterior distribution of the position of the $n$-th tracked object. Given~(\ref{eq:optimal_qs}) and the DVAE generative and inference models, we rapidly see that at a given time $t$, the distribution over $\mathbf{s}_{tn}$ has non-linear dependencies w.r.t.\ the previous and current DVAE latent variables $\mathbf{z}_{1:t,n}$ and the previous source positions $\mathbf{s}_{1:t-1,n}$. These non-linear dependencies impede to obtain an efficient closed-form solution. We resort to point sample estimates obtained using samples of $\mathbf{z}_{1:t,n}$ and of $\mathbf{s}_{1:t-1,n}$, at the current iteration $\mathbf{z}_{1:t,n}^{(i)}$ and $\mathbf{s}_{1:t-1,n}^{(i)}$. With these samples the posterior distribution is approximated with (the details can be found in Appendix~\ref{appe:E-S-step}):
\begin{equation}
    q_{\phi_{\mathbf{s}}}(\mathbf{s}_{:, n}|\mathbf{o}) \approx \prod_{t=1}^T q_{\phi_{\mathbf{s}}}(\mathbf{s}_{tn} | \mathbf{s}_{1:t-1, n}^{(i)}, \mathbf{z}_{1:t, n}^{(i)}, \mathbf{o}),\label{eq:optimal-qs-fact-over-t}
\end{equation}
where each term of the product in is shown to be a Gaussian:
\begin{equation}
q_{\phi_{\mathbf{s}}}(\mathbf{s}_{tn}| \mathbf{s}_{1:t-1, n}^{(i)}, \mathbf{z}_{1:t, n}^{(i)}, \mathbf{o}) = \mathcal{N}(\mathbf{s}_{tn} ; \mathbf{m}_{tn}, \mathbf{V}_{tn}),
\label{eq:posterior-s-Gaussian}
\end{equation}
with covariance matrix and mean vector given by:
\begin{equation}
    \mathbf{V}_{tn} = \Big(\textstyle \sum\limits_{k=1}^{K_t} \eta_{tkn} \boldsymbol{\Phi}_{tk}^{-1}+ \textrm{diag} (\boldsymbol{v}^{(i)}_{\theta_{\mathbf{s}},tn})^{-1}\Big)^{-1},
    \label{eq:posterior-s-cov}
\end{equation}
\begin{equation} 
    \mathbf{m}_{tn} = \mathbf{V}_{tn} \Big(\textstyle \sum\limits_{k=1}^{K_t} \eta_{tkn} \boldsymbol{\Phi}_{tk}^{-1} \mathbf{o}_{tk} + \textrm{diag} (\boldsymbol{v}^{(i)}_{\theta_{\mathbf{s}},tn})^{-1} \boldsymbol{\mu}^{(i)}_{\theta_{\mathbf{s}},tn}\Big),\label{eq:posterior-s-mean}
\end{equation}
where $\boldsymbol{v}^{(i)}_{\theta_{\mathbf{s}},tn}$ and $\boldsymbol{\mu}^{(i)}_{\theta_{\mathbf{s}},tn}$ are simplified notations for $\boldsymbol{v}_{\theta_{\mathbf{s}}}(\mathbf{s}_{1:t-1, n}^{(i)}, \mathbf{z}_{1:t, n}^{(i)})$ and $\boldsymbol{\mu}_{\theta_{\mathbf{s}}}(\mathbf{s}_{1:t-1, n}^{(i)}, \mathbf{z}_{1:t, n}^{(i)})$, respectively denoting the variance and mean vector provided by the DVAE decoder network for object $n$ at time frame $t$. As we have to sample both $\mathbf{s}_{:,n}$ and $\mathbf{z}_{:,n}$, we need to pay attention to the sampling order. This will be discussed in detail in Section~\ref{subsec:sampling-order}. Importantly, in practice, $\mathbf{m}_{tn}$ is used as the estimate of $\mathbf{s}_{tn}$.


Eq. \eqref{eq:posterior-s-mean} shows that the estimated position vector for object $n$ is obtained by combining the observations $\mathbf{o}_{tk}$, i.e., the detected object positions, and the mean position vector $\boldsymbol{\mu}^{(i)}_{\theta_{\mathbf{s}},tn}$ predicted by the DVAE generative model. The balance between these two terms depends on the assignment variables $\eta_{tkn}$, the observation model covariance matrix $\boldsymbol{\Phi}_{tk}$ and the position vector covariance matrix predicted by the DVAE generative model $\boldsymbol{v}_{\theta_{\mathbf{s}},tn}^{(i)}$. Ideally, the model should be able to appropriately balance the importance of these two terms so as to optimally exploit both the observations and  the DVAE predictions. 

\subsubsection{E-Z step}
\label{sssec:E-Z-step}
\noindent In the E-Z step, we consider the posterior distribution $q_{\phi_{\mathbf{z}}}(\mathbf{z} | \mathbf{s})$ of the DVAE. This distribution is defined by \eqref{eq:qz_factorises}, \eqref{eq:DVAE-inference-model-chained} and \eqref{eq:DVAE-inference-model-Gaussian}.
 In \eqref{eq:elbo}, the corresponding term is the third one, which we denote by $ \mathcal{L}_{\mathbf{z}}(\theta_{\mathbf{s}}, \theta_{\mathbf{z}}, \phi_{\mathbf{z}}; \mathbf{o})$ and which factorises across objects as follows (see Appendix~\ref{appe:E-Z-VLB}):
\begin{equation}
     \mathcal{L}_{\mathbf{z}}(\theta_{\mathbf{s}}, \theta_{\mathbf{z}}, \phi_{\mathbf{z}}; \mathbf{o}) = \sum_{n=1}^N \mathcal{L}_{\mathbf{z},n}(\theta_{\mathbf{s}}, \theta_{\mathbf{z}}, \phi_{\mathbf{z}}; \mathbf{o}),
     \label{eq:DVAE-VLB-in-MOT}
\end{equation}
with
\begin{align}
    & \mathcal{L}_{\mathbf{z},n}(\theta_{\mathbf{s}}, \theta_{\mathbf{z}}, \phi_{\mathbf{z}}; \mathbf{o}) \nonumber
    \\& \qquad = \mathbb{E}_{ q_{\phi_{\mathbf{s}}}(\mathbf{s}_{:,n}|\mathbf{o})} \Big[ \mathbb{E}_{q_{\phi_{\mathbf{z}}}(\mathbf{z}_{:,n} | \mathbf{s}_{:,n})} \big[\log p_{\theta_{\mathbf{s}\mathbf{z}}}(\mathbf{s}_{:,n}, \mathbf{z}_{:,n})\big] \nonumber \\ & \qquad - \mathbb{E}_{ q_{\phi_{\mathbf{z}}}(\mathbf{z}_{:,n} | \mathbf{s}_{:,n})} \big[\log q_{\phi_{\mathbf{z}}}(\mathbf{z}_{:,n} | \mathbf{s}_{:,n})\big] \Big]. \label{eq:DVAE-VLB-in-MOT-n}
\end{align}
Inside the expectation $\mathbb{E}_{ q_{\phi_{\mathbf{s}}}(\mathbf{s}_{:,n}|\mathbf{o})} [\cdot]$, we recognize the ELBO of the DVAE model defined in \eqref{eq:DVAE-ELBO-general-form-a} and applied to object $n$. This suggests the following strategy. Previously to and independently of the DVAE-UMOT algorithm, as briefly stated before, we pre-train the DVAE model on a dataset of synthetic single-object unlabeled sequences (this is detailed in Section~\ref{subsec:DVAE-pre-training}). This is done only once, and the resulting DVAE is then plugged into the DVAE-UMOT algorithm to process multi-object sequences. This provides the E-Z step with very good initial values of the DVAE parameters $\theta_{\mathbf{s}}$, $\theta_{\mathbf{z}}$ and $\phi_{\mathbf{z}}$. As for the following of the E-Z step, the expectation over $q_{\phi_{\mathbf{s}}}(\mathbf{s}_{:,n}|\mathbf{o})$ in \eqref{eq:DVAE-VLB-in-MOT-n} is not analytically tractable. A Monte Carlo estimate is thus used instead, using samples of both $\mathbf{z}$ and $\mathbf{s}$, similarly to what was done in the E-S step. Finally, SGD is used to maximize (the Monte Carlo estimate of) $\mathcal{L}_{\mathbf{z}}(\theta_{\mathbf{s}}, \theta_{\mathbf{z}}, \phi_{\mathbf{z}}; \mathbf{o})$, jointly updating $\theta_{\mathbf{s}}$, $\theta_{\mathbf{z}}$ and $\phi_{\mathbf{z}}$; that is, we fine-tune the DVAE model within the DVAE-UMOT algorithm, using the observations $\mathbf{o}$. Note that in our experiments, we also consider the case where we neutralize the fine-tuning, i.e., we remove the E-Z step and use the DVAE model as provided by the pre-training phase.

\subsubsection{E-W step}
\label{sssec:E-W-step}

\noindent Thanks to the separation of $\mathbf{w}$ from the two other latent variables in \eqref{eq:inf_approximation}, the posterior distribution $q_{\phi_{\mathbf{w}}}(\mathbf{w} | \mathbf{o})$ can be calculated in closed form by directly applying the optimal structured mean-field update equation~\eqref{eq:optimal-factorized-posterior-general} to our model. It can be shown that this is equivalent to maximizing \eqref{eq:elbo} w.r.t.\ $q_{\phi_{\mathbf{w}}}(\mathbf{w} | \mathbf{o})$. We obtain (see Appendix~\ref{appe:VEM-calculations} for details):
\begin{align}
    q_{\phi_{\mathbf{w}}}(\mathbf{w} | \mathbf{o})
    & \propto \prod_{t=1}^T \prod_{k=1}^{K_t} q_{\phi_{\mathbf{w}}}(w_{tk}|\mathbf{o}), \label{e70}
\end{align}
with 
\begin{equation}
    q_{\phi_{\mathbf{w}}}(w_{tk} = n|\mathbf{o}) = \eta_{tkn} = \frac{\beta_{tkn}}{\sum_{i=1}^N \beta_{tki}},\label{eq:posterior-W-eta}
\end{equation}
where
\begin{equation}
    \beta_{tkn} = \mathcal{N}(\mathbf{o}_{tk} ;  \mathbf{m}_{tn}, \boldsymbol{\Phi}_{tk})\exp \Big(-\frac{1}{2}\text{Tr}\big( \boldsymbol{\Phi}^{-1}_{tk} \mathbf{V}_{tn}\big)\Big), \label{eq:posterior-W-beta}
\end{equation}
where $\mathbf{m}_{tn}$ and $\mathbf{V}_{tn}$ are defined in~\eqref{eq:posterior-s-mean} and \eqref{eq:posterior-s-cov}, respectively.

\subsubsection{M step}
\noindent As discussed in Section~\ref{sec:Background}, the maximization step generally consists in estimating the parameters $\theta$ of the generative model by maximizing the ELBO over $\theta$. We recall that $\theta = \{\theta_{\mathbf{o}} = \{\boldsymbol{\Phi}_{tk}\}_{t,k=1}^{T, K_t}, \theta_{\mathbf{s}}, \theta_{\mathbf{z}}\}$. 
In this work, the parameters of the DVAE decoder $\theta_{\mathbf{s}}$ and $\theta_{\mathbf{z}}$ are first  estimated (offline) by the pre-training of the DVAE and then fine-tuned in the E-Z step, all this jointly with the parameters of the encoder $\phi_{\mathbf{z}}$. Therefore, in the M-step, we only need to estimate the observation model covariance matrices $\theta_{\mathbf{o}} = \{\boldsymbol{\Phi}_{tk}\}_{t,k=1}^{T, K_t}$. 
In \eqref{eq:elbo}, only the first term depends on $\theta_{\mathbf{o}}$. Setting its derivative with respect to $\boldsymbol{\Phi}_{tk}$ to zero, we obtain (see Appendix~\ref{appe:VEM-calculations} for details):
\begin{align}
     \boldsymbol{\Phi}_{tk} &= \sum_{n=1}^N \eta_{tkn}\Big((\mathbf{o}_{tk} - \mathbf{m}_{tn})(\mathbf{o}_{tk} - \mathbf{m}_{tn})^T 
     + \mathbf{V}_{tn}\Big).
     \label{eq:obs-cov-matrix-update}
\end{align}
In practice, it is difficult to obtain a reliable estimation using only a single observation. We address this issue in Section~\ref{subsec:observation-variance}.

\section{Algorithm Implementation}


\subsection{DVAE-UMOT initialization}\label{subsec:MOT-initialization}

\begin{algorithm}[!t]
\renewcommand{\algorithmicrequire}{\textbf{Input:}}
\renewcommand{\algorithmicensure}{\textbf{Output:}}
\algsetblock[Name]{Initialization}{Stop}{1}{0.5cm}
\caption{DVAE-UMOT algorithm}\label{algo1}
\begin{algorithmic}[1]
\Require 
\Statex Detected bounding boxes $\mathbf{o}= \mathbf{o}_{1:T, 1:K_t}$;
\Ensure 
\Statex Parameters of  $q_{\phi_{\mathbf{s}}}(\mathbf{s}): \{\mathbf{m}_{tn}, \mathbf{V}_{tn}\}_{t, n=1}^{T, N}$ (the estimated position of each tracked object $n$ at each time frame $t$ is $\mathbf{m}_{tn}$);  
\Statex Values of the assignment variable $\{\eta_{tkn}\}_{t,n, k=1}^{T, N, K_t}$;
\Initialization
    \State See Section~\ref{subsec:MOT-initialization}
\For{$i \leftarrow 1$ to $I$}
    
    \State \textbf{E-W Step}
    
    \For{$n \leftarrow 1$ to $N$}
        \For{$t \leftarrow 1$ to $T$}
        \For{$k \leftarrow 1$ to $K_t$}
        \State Compute $\eta^{(i)}_{tkn}$ using  (\ref{eq:posterior-W-eta}) and  (\ref{eq:posterior-W-beta});
        \EndFor
        \EndFor
    \EndFor

    \State \textbf{E-Z and E-S Step}
    \For{$n \leftarrow 1$ to $N$}
        \For{$t \leftarrow 1$ to $T$}
        \State 
        \textbf{\textit{Encoder}};
        \State \multiline{Compute $\boldsymbol{\mu}^{(i)}_{\phi_{\mathbf{z}}, tn}$, $\boldsymbol{v}^{(i)}_{\phi_{\mathbf{z}}, tn}$ with input $\mathbf{s}^{(i-1)}_{1:t, n}$ and $\mathbf{z}^{(i)}_{1:t-1,n}$;}
        \State \multiline{Sample $\mathbf{z}^{(i)}_{tn}$ from $q_{\phi_{\mathbf{z}}}(\mathbf{z}_{tn} | \mathbf{s}_{1:t, n}^{(i-1)}, \mathbf{z}_{1:t-1, n}^{(i)}) = \mathcal{N}\big(\mathbf{z}_{tn} ; \boldsymbol{\mu}^{(i)}_{\phi_{\mathbf{z}},tn}, \textrm{diag}(\boldsymbol{v}^{(i)}_{\phi_{\mathbf{z}},tn})\big)$;}
     
        \State \textbf{\textit{Decoder}};
        \State \multiline{Compute $\boldsymbol{\mu}^{(i)}_{\theta_{\mathbf{z}},tn}$ and $\boldsymbol{v}^{(i)}_{\theta_{\mathbf{z}},tn}$ with input $\mathbf{s}^{(i)}_{1:t-1, n}$ and $\mathbf{z}^{(i)}_{1:t-1, n}$;}

        \State \multiline{Compute $\boldsymbol{\mu}^{(i)}_{\theta_{\mathbf{s}},tn}$ and $\boldsymbol{v}^{(i)}_{\theta_{\mathbf{s}},tn}$ with input $\mathbf{s}^{(i)}_{1:t-1, n}$ and $\mathbf{z}^{(i)}_{1:t, n}$;}
        \State \textbf{\textit{E-S update}};
        \State \multiline{Compute $\mathbf{m}^{(i)}_{tn}$, $\mathbf{V}^{(i)}_{tn}$ using  (\ref{eq:posterior-s-mean}) and (\ref{eq:posterior-s-cov});}
        \State \multiline{Sample $\mathbf{s}^{(i)}_{tn}$ from $\mathcal{N}(\mathbf{s}_{tn} ; \mathbf{m}^{(i)}_{tn}, \mathbf{V}^{(i)}_{tn})$;}
        \EndFor
        \State \textbf{\textit{E-Z update}};
        \State \multiline{Compute $\widehat{\mathcal{L}}_{n}(\theta_{\mathbf{s}}, \theta_{\mathbf{z}}, \phi_{\mathbf{z}}; \mathbf{o})$ using  (\ref{eq:DVAE-VLB-in-MOT-n-estimate});}
    \EndFor
    \State \multiline{Compute $\widehat{\mathcal{L}}(\theta_{\mathbf{s}}, \theta_{\mathbf{z}}, \phi_{\mathbf{z}}; \mathbf{o}) = \sum_{n=1}^N \widehat{\mathcal{L}}_{n}(\theta_{\mathbf{s}}, \theta_{\mathbf{z}}, \phi_{\mathbf{z}}; \mathbf{o})$;}
    \State \multiline{Fine-tune the DVAE parameters \{$\theta_{\mathbf{s}}$, $\theta_{\mathbf{z}}$, $\phi_{\mathbf{z}}$\} by applying SGD on $\widehat{\mathcal{L}}(\theta_{\mathbf{s}}, \theta_{\mathbf{z}}, \phi_{\mathbf{z}}; \mathbf{o})$;}
    \State \textbf{M Step}
    \State \multiline{Compute $\boldsymbol{\Phi}^{(i)}_{tk}$ using (\ref{eq:obs-cov-matrix-update}) or following Section~\ref{subsec:observation-variance};}
\EndFor
\end{algorithmic}
\end{algorithm}

Before starting the EM iterations of the proposed DVAE-UMOT algorithm, we need to initialize the values of several parameters and variables. Theoretically, there is no preference in the order of the three E-steps. In practice however, we chose to follow the order E-W Step, E-Z Step, E-S Step, and finally M Step, for initialization convenience. Indeed, starting with the M Step or the E-S/E-Z steps, requires the initialization of the assignment variables $\mathbf{w}_{1:T,1:K_t}$, which poses a problem because we do not have any prior knowledge about them. Instead, starting with E-W requires the initialisation of the object bounding boxes and their covariance matrices through the sequence, as well as the observation covariance matrices.
 
 
The estimate of $\mathbf{s}_{tn}$, $\mathbf{m}_{tn}$, can be easily initialised over a short sequence by assuming that the object does not move too much. Indeed, the initial values of $\mathbf{m}_{tn}$ can be set to the value of the observed bounding box at the beginning of the sequence $\mathbf{m}_{0n}$. While this strategy is very straightforward to implement, it is too simple for many tracking scenarios, especially for long sequences. We thus propose to split a long sequence into sub-sequences. For each sub-sequence, we initialise $\mathbf{m}_{tn}$ to the value at the beginning of the sub-sequence. After this initialisation, we run a few iterations of DVAE-UMOT over the sub-sequence, allowing us to have an estimate of the object bounding box at the end of the sub-sequence. This value will then be used to provide a constant initialisation for the next sub-sequence. At the end, all these initializations are concatenated, thus providing a piece-wise constant initialization for DVAE-UMOT over the entire long sequence. After computing the E-W step, a sequence of samples from the previous iteration is sent through the E-S/E-Z steps, and the same initialisation values are taken. The implementation details as well as the pseudo-code of this cascade initialization strategy are provided in Appendix~\ref{appe:cascade-init}.

 
Finally, for the initialization of the observation model covariance matrices $\boldsymbol{\Phi}_{tk}$, see Section \ref{subsec:observation-variance}. The covariance matrices of the object bounding boxes $\mathbf{V}_{tn}$ are initialized with the same values as $\boldsymbol{\Phi}_{tk}$.

\subsection{Sampling order}
\label{subsec:sampling-order}

As already mentioned in Section~\ref{sec:E-S}, we must pay attention to the sampling order of $\mathbf{s}$ and $\mathbf{z}$ when running the iterations of the E-S and E-Z steps. As indicated in the pseudo-code of Algorithm~\ref{algo1}, in practice, the E-S and E-Z steps are processed jointly. We start with the initial position vector sequence $\mathbf{s}^{\text{(0)}}_{1:T,1:N}$ and the initial mean position vector sequence $\mathbf{m}^{\text{(0)}}_{1:T,1:N}$. At any iteration $i$ of the E-Z and E-S steps, for each tracked object $n$ and each time frame $t$, we sample in the following order:
\begin{enumerate}[leftmargin=0.45cm]
    \item Compute the parameters  $\boldsymbol{\mu}^{(i)}_{\phi_{\mathbf{z}}, tn}$ and $\boldsymbol{v}^{(i)}_{\phi_{\mathbf{z}}, tn}$\footnote{$\boldsymbol{\mu}^{(i)}_{\phi_{\mathbf{z}}, tn}$ and $\boldsymbol{v}^{(i)}_{\phi_{\mathbf{z}}, tn}$ are shortcuts for $\boldsymbol{\mu}_{\phi_{\mathbf{z}}}\big(\mathbf{s}_{1:t, n}^{(i-1)}, \mathbf{z}_{1:t-1, n}^{(i)}\big)$ and $\boldsymbol{v}_{\phi_{\mathbf{z}}}\big(\mathbf{s}_{1:t, n}^{(i-1)}, \mathbf{z}_{1:t-1, n}^{(i)}\big)$ respectively.} of the posterior distribution of $\mathbf{z}_t$ using the DVAE encoder network with inputs $\mathbf{s}^{(i-1)}_{1:t,n}$ sampled at the previous iteration and $\mathbf{z}_{1:t-1, n}^{(i)}$ sampled at the current iteration. Then, sample $\mathbf{z}_{tn}^{(i)}$ from  $q_{\phi_{\mathbf{z}}}(\mathbf{z}_{tn} | \mathbf{s}_{1:t, n}^{(i-1)}, \mathbf{z}_{1:t-1, n}^{(i)})$.
    \item Compute the parameters  $\boldsymbol{\mu}^{(i)}_{\theta_{\mathbf{z}}, tn}$ and $\boldsymbol{v}^{(i)}_{\theta_{\mathbf{z}}, tn}$\footnote{Analogous definitions hold.} of the generative distribution of $\mathbf{z}_t$ using the corresponding DVAE network with inputs $\mathbf{s}^{(i)}_{1:t-1,n}$ and $\mathbf{z}_{1:t-1, n}^{(i)}$, both sampled at the current iteration.
    \item Compute the parameters  $\boldsymbol{\mu}^{(i)}_{\theta_{\mathbf{s}}, tn}$ and $\boldsymbol{v}^{(i)}_{\theta_{\mathbf{s}}, tn}$ of the generative distribution of $\mathbf{s}_t$ using the corresponding DVAE decoder network with inputs $\mathbf{s}^{(i)}_{1:{t-1}, n}$  and $\mathbf{z}_{1:t, n}^{(i)}$, both sampled at the current iteration. Compute the parameters $\mathbf{m}^{(i)}_{tn}$ and  $\mathbf{V}^{(i)}_{tn}$ of the posterior distribution of $\mathbf{s}_t$ using \eqref{eq:posterior-s-cov} and \eqref{eq:posterior-s-mean}, and sample $\mathbf{s}^{(i)}_{tn}$ from it.
\end{enumerate}
Note that with the above sampling order, the Monte Carlo estimate of the ELBO term maximized in the E-Z step \eqref{eq:DVAE-VLB-in-MOT-n} is given by (for object $n$):
\begin{align}
    & \widehat{\mathcal{L}}_{\mathbf{z},n}(\theta_{\mathbf{s}}, \theta_{\mathbf{z}}, \phi_{\mathbf{z}}; \mathbf{o}) = \sum_{t=1}^T \log p_{\theta_{\mathbf{s}}}(\mathbf{s}^{(i)}_{tn} | \mathbf{s}^{(i)}_{1:t-1, n}, \mathbf{z}^{(i)}_{1:t, n}) \nonumber \\
    & - \sum_{t=1}^T D_{\textsc{kl}}\big(q_{\phi_{\mathbf{z}}}(\mathbf{z}_{tn} | \mathbf{s}^{(i-1)}_{1:t, n}, \mathbf{z}^{(i)}_{1:t-1, n}) || p_{\theta_{\mathbf{z}}}(\mathbf{z}_{tn} | \mathbf{s}^{(i)}_{1:t-1,n}, \mathbf{z}^{(i)}_{1:t-1,n})\big).\label{eq:DVAE-VLB-in-MOT-n-estimate}
\end{align}
The whole DVAE-UMOT algorithm, taking into account these practical aspects, is summarized in the form of pseudo-code in Algorithm~\ref{algo1}.

\subsection{Variance estimation}\label{subsec:observation-variance}

In our experiments, we observed that the estimated values of both $\boldsymbol{\Phi}_{tk}$ and $\boldsymbol{v}_{\theta_{\mathbf{s}},tn}$ in \eqref{eq:posterior-s-mean} increased very quickly with the DVAE-UMOT algorithm iterations. This caused instability and unbalance between these two terms, which finally conducted the whole model to diverge. To solve this problem, we set $\boldsymbol{\Phi}_{tk}$ to a given fixed value, which is constant on the whole analyzed $T$-frame sequence and not updated during the DVAE-UMOT iterations. Specifically, the diagonal of $\boldsymbol{\Phi}_{tk}$ is set to $r_{\boldsymbol{\Phi}}^2\big[(o_{1k}^R-o_{1k}^L)^2,(o_{1k}^T-o_{1k}^B)^2,(o_{1k}^R-o_{1k}^L)^2,(o_{1k}^T-o_{1k}^B)^2\big]$, where $r_{\boldsymbol{\Phi}}$ is a factor lower than 1. In common terms, $\boldsymbol{\Phi}_{tk}$ is set to a fraction of the (squared) size of the corresponding observation at frame 1. This turned out to stabilise the iteration process and finally led to very satisfying tracking results. The value of $r_{\boldsymbol{\Phi}}$ is a hyperparameter of the model and its influence is experimentally evaluated in Section~\ref{subsec:ablation-study}.

\section{Experiments}
\label{sec:experiments}

\subsection{Choice of the DVAE model}\label{subsec:DVAE-model}

\par We recall that the DVAE is a general class of models that differ by adopting different conditional independence assumptions for the generative distributions in the right-hand-side of (\ref{eq:DVAE-joint-chained}). Seven DVAE models have been extensively discussed, and six of them have been benchmarked in the analysis-resynthesis task (on speech signals and 3D human motion data) in \cite{MAL-089}. We chose to use here the stochastic recurrent neural network (SRNN) model initially proposed in \cite{fraccaro2016sequential}, because it was shown to provide a very good trade-off between complexity and modeling power in  \cite{MAL-089}. The probabilistic dependencies of the SRNN generative model are defined as follows:
\begin{equation}
    p_{\theta_{\mathbf{sz}}}(\mathbf{s}_{1:T}, \mathbf{z}_{1:T}) = \prod_{t=1}^Tp_{\theta_{\mathbf{s}}}(\mathbf{s}_t|\mathbf{s}_{1:t-1}, \mathbf{z}_{t})p_{\theta_{\mathbf{z}}}(\mathbf{z}_t|\mathbf{s}_{1:t-1}, \mathbf{z}_{t-1}).\label{eq:SRNN-generative-model}
\end{equation}
To perform online tracking, we propose to use the following causal SRNN inference model:
\begin{equation}
    q_{\phi_{\mathbf{z}}}(\mathbf{z}_{1:T}| \mathbf{s}_{1:T}) = \prod_{t=1}^T q_{\phi_{\mathbf{z}}}(\mathbf{z}_t|\mathbf{s}_{1:t}, \mathbf{z}_{t-1}).\label{eq:SRNN-inference-model-causal}
\end{equation}

The SRNN generative model is implemented with a forward LSTM network, which embeds all the past information of the sequence $\mathbf{s}$. Then, a dense layer with the tanh activation function plus a linear layer provide the parameters $\boldsymbol{\mu}_{\theta_{\mathbf{s}}}, \boldsymbol{v}_{\theta_{\mathbf{s}}}$.  Similarly, the parameters $\boldsymbol{\mu}_{\theta_{\mathbf{z}}}, \boldsymbol{v}_{\theta_{\mathbf{z}}}$ are computed with two dense layers with tanh activation function plus a linear layer appended to the LSTM as well. The inference model shares the hidden variables of the forward LSTM network of the generative model and uses two dense layers with the tanh activation function plus a linear layer to compute the parameters $\boldsymbol{\mu}_{\phi_{\mathbf{z}}}, \boldsymbol{v}_{\phi_{\mathbf{z}}}$. More implementation details can be found in Appendix~\ref{appe:srnn-implementation}.


\subsection{DVAE pre-training}
\label{subsec:DVAE-pre-training}

\subsubsection{Dataset}
\label{sssec:synthetic-dataset}

We generated synthetic bounding box trajectories, in the form of $T$-frame sequences ($T=60$) of 4D vectors $\{(x_t^L, x_t^T, x_t^R, x_t^B)\}_{t=1}^T$ gathering the top-left and bottom-right bounding box 2D coordinates. To generate bounding boxes with reasonable size, we do not generate the four dimensions separately. Instead, we  generate the coordinates of the top-left point plus the height and width of the bounding boxes and deduce the coordinates of the bottom-right point. The width-height ratio is sampled randomly, and kept constant during the trajectory. The other three values (top-left coordinates and width) are generated using piece-wise combinations of several elementary dynamic functions, namely: static $a(t)=a_0$, constant velocity $a(t)=a_1t+a_0$, constant acceleration $a(t)=a_2t^2+a_1t+a_0$, and sinusoidal (allowing for circular trajectories) $a(t)=a\sin(\omega t+\phi_0)$. An example of a 3-segment combination could be:
\begin{equation}
 a(t) = \left\{\begin{array}{ll}
a_0 & 1\leq t<t_1,\\
a_2 t^2 + a_1 t + a_0' & t_1\leq t < t_2, \\
a_3\sin(\omega t+\phi_0) & t_1\leq t \leq T,
 \end{array}
\right.
\end{equation}
where the segments length is sampled from some pre-defined distributions to generate reasonable and continuous trajectories. The parameters $a_1$, $a_2$, $\omega$, and $\phi_0$ are also sampled from some pre-defined distributions whose parameters are estimated from the detections on the MOT17 training dataset published in \cite{dendorfer2020motchallenge}. The two remaining parameters, $a_0$ and $a$, are set to the values that ensure continuous trajectories. More technical details about the synthetic trajectories generation can be found in Appendix~\ref{appe:synthetic-dataset}.

\begin{figure}[!t]
    \centering
    \subfloat[Short sequence dataset\label{fig4:sub1}]{\includegraphics[width=.49\linewidth]{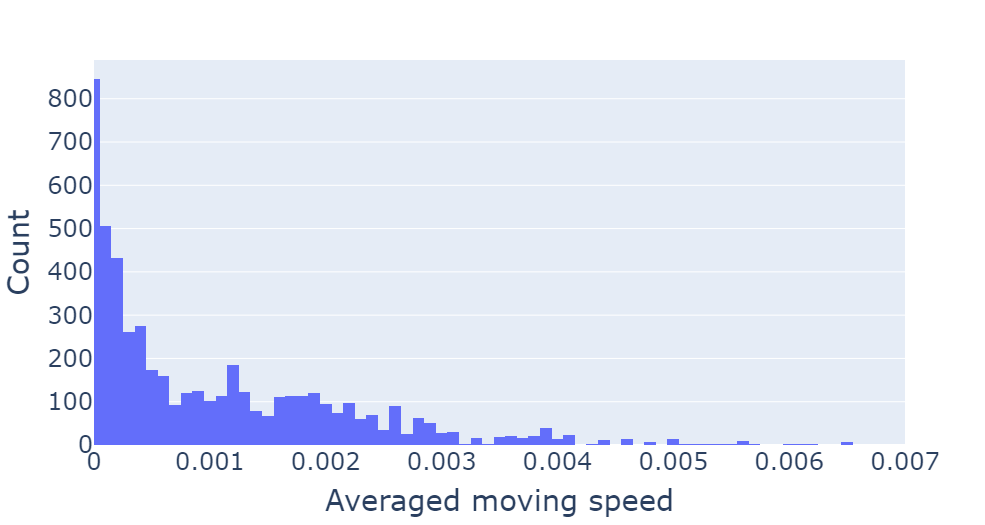}}
    \hfill
    \subfloat[Medium sequence dataset\label{fig4:sub2}]{\includegraphics[width=.49\linewidth]{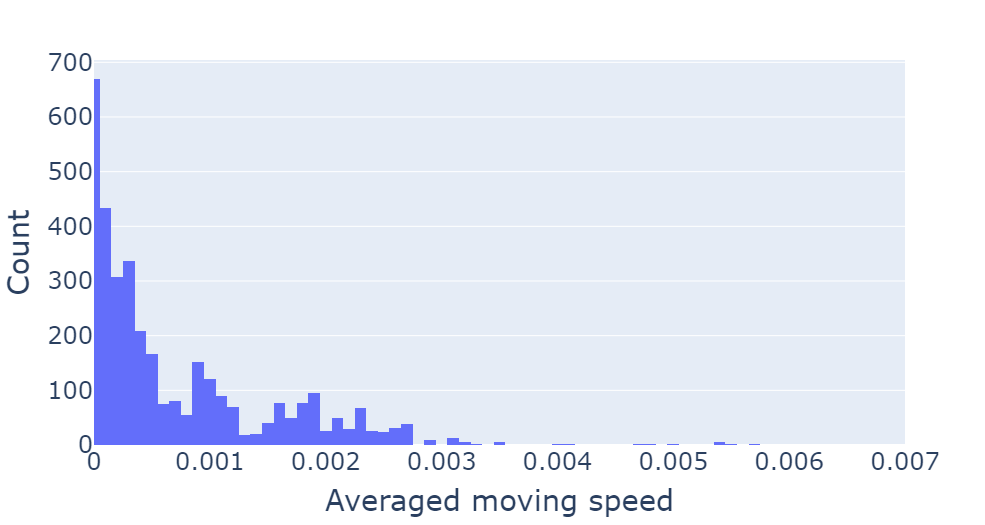}}
    \\
    \vskip -0.25cm
    \subfloat[Long sequence dataset\label{fig4:sub3}]{\includegraphics[width=.49\linewidth]{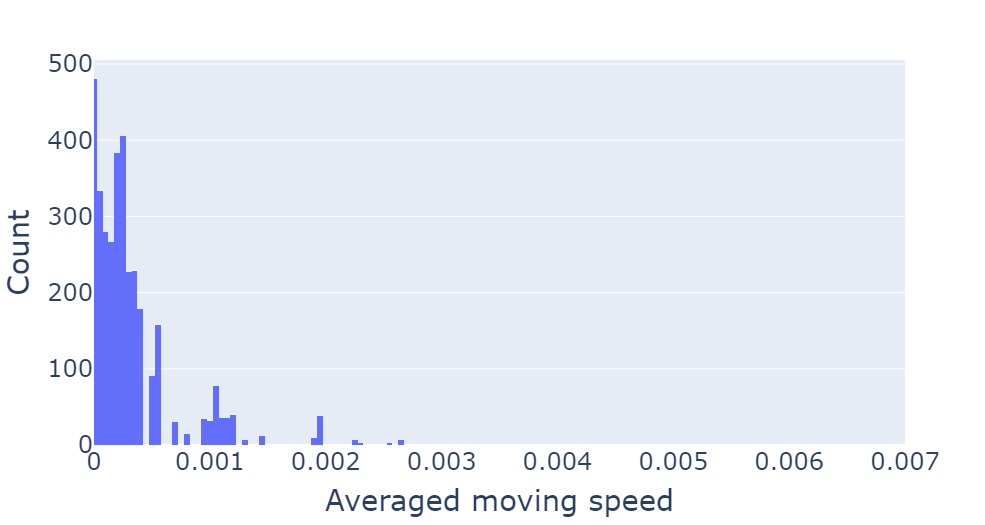}}
    \hfill
    \subfloat[Pre-training dataset\label{fig4:sub4}]{\includegraphics[width=.49\linewidth]{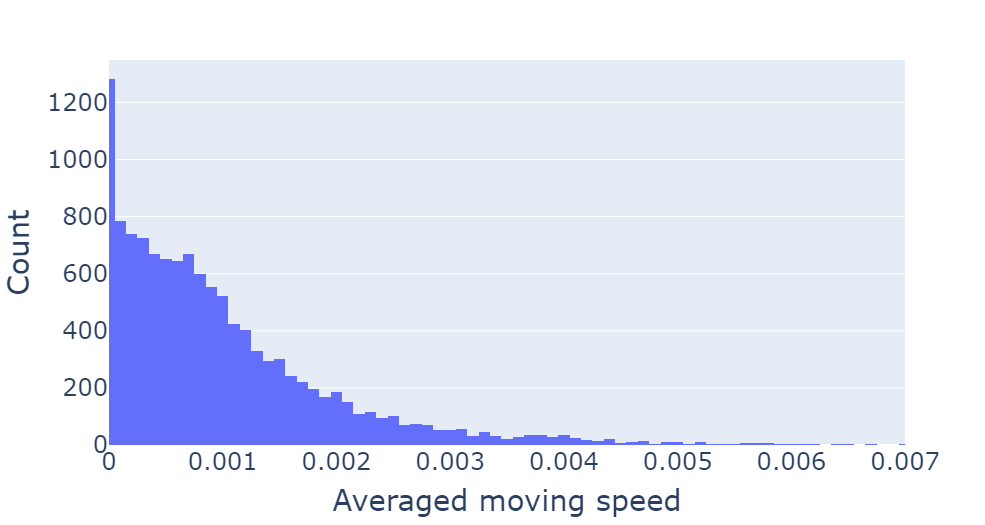}}
\caption{Histograms of the averaged speed for each tracking sequence in the four datasets.}
\label{fig:figure4}
\end{figure}

In \figurename~\ref{fig4:sub4}, we show the histogram of the average velocity computed for each sequence of the synthetic dataset. We observe that the dynamics of the synthetic dataset cover a wide range of velocities. 
Finally, we have generated $12,105$ sequences for the training dataset and $3,052$ sequences for the validation dataset.

\subsubsection{Training details}
\label{sssec:training-details}

The SRNN model used in our experiments is an auto-regressive model, i.e., it uses the past observations $\mathbf{s}_{1:t-1}$ to predict the present one $\mathbf{s}_{t}$. We trained the model in teacher-forcing mode \cite{williams1989learning}. This means that during training, we used the ground-truth past observations to generate the current one, and not the previously generated past observations (see \cite[Chapter 4]{MAL-089} for a discussion on this issue). The model was trained using the Adam optimizer with a learning rate set to $0.001$ and a batch size set to $256$. An early-stopping strategy was adopted, with a patience of $50$ epochs.

\subsection{DVAE-UMOT evaluation set-up} 


\subsubsection{Dataset} \label{sssec:MOT-dataset}
\begin{table}[!t]
\centering
\caption{Key Characteristics of the MOT17 Training Set}
\label{table:MOT17}
 \begin{tabularx}{0.485\textwidth}{ccccccc}

 \toprule
 & & \# frames &  \multicolumn{4}{c}{\# tracks longer than $T$ frames}\\
 Seq. & FPS & (duration) & $T=0$ & $T=60$ & $T=120$ & $T=300$ \\
 \midrule
 02 & 30 & 600 (20s) & 62 & 51 & 45 & 32 \\
 04 & 30 & 1050 (35s) & 83 & 77 & 72 & 57 \\
 05 & 14 & 837 (60s) & 133 & 29 & 15 & 3 \\
 09 & 30 & 525 (18s) & 26 & 22 & 17 & 5 \\
 10 & 30 & 654 (22s) & 57 & 48 & 34 & 14 \\
 11 & 30 & 900 (30s) & 75 & 43 & 22 & 5\\
 13 & 25 & 750 (30s) & 110 & 71 & 42 & 3 \\
 \bottomrule
 \end{tabularx}
\end{table}

For the evaluation of the proposed DVAE-UMOT algorithm, we used the training set of the MOT17 dataset. MOT17 is a widely used pedestrian tracking dataset. It contains pedestrian tracking sequences filmed on different scenes such as in a shopping mall or in a street, with static or moving cameras. The motion patterns of the pedestrians in these videos are quite diverse and challenging. Table~\ref{table:MOT17} shows a summary of the MOT17 training set features. This training set contains in total seven sequences with length varying from twenty seconds to one minute. Five of them have a frame rate of 30 fps while that of the other two sequences is 14 and 25 fps, respectively. The ground-truth bounding boxes are provided, as well as the detection results of three customized detectors, namely DPM \cite{5255236}, Faster-RCNN \cite{NIPS2015_14bfa6bb}, and SDP \cite{7780603}.

\par As briefly stated in the introduction, we focus our study on the analysis of a scene with a fixed number of tracks, without considering the tracks birth/death process. So, we cannot use the MOT17 dataset as it is. We have thus designed a new dataset from the MOT17 training set, which we call the MOT17-3T dataset. First, we matched the detected bounding boxes to the ground-truth bounding boxes using the Hungarian algorithm \cite{kuhn1955hungarian} and retained only the matched detected bounding boxes (i.e., the detected bounding boxes that were not matched to any ground-truth bounding boxes were discarded). The cost matrix were computed according to the the Intersection-over-Union (IoU) distance between bounding boxes. We split each complete video sequence into subsequences of length $T$ (three different values of $T$ are tested in our experiments, as detailed below) and only kept the tracks with a length no shorter than $T$. For each subsequence, we randomly chose three tracks that appeared in this subsequence from the beginning to the end. The detected bounding boxes of these three tracks form one test data sample. We have tested three values for the sequence length $T$ to evaluate its influence on the tracking performance of our algorithm: 60, 120, and 300 frames (respectively corresponding to 2, 4, and 10 seconds at 30 fps). Some statistics of the tracks can be found in Table~\ref{table:MOT17}. Among the three public detection results provided with the MOT17 dataset, SDP has the best detection performance. So, we used the detection results of SDP to create our dataset. We have finally created $1{,}712$, $1{,}161$, and $1{,}137$ multi-track test sequences of length $T= 60, 120$ and $300$ frames, respectively.

\subsubsection{Algorithm settings}

For the DVAE-UMOT algorithm, there are four hyperparameters to be set. The observation covariance matrix ratio $r_{\boldsymbol{\Phi}}$ is set to $0.04$, the initialization subsequence length $J$ is set to 30, and the initialization iteration number $I_0$ is set to 20. The DVAE-UMOT algorithm itself is run for $I = 70$ iterations, which was experimentally shown to lead to convergence.

\subsubsection{Baselines}

We compare our model with two very recent state-of-the-art probabilistic MOT methods: The Autoregressive Tracklet Inpainting and Scoring for Tracking (ArTIST) model \cite{saleh2021probabilistic} and the Variational Kalman Filter (VKF) \cite{8907431}. 

ArTIST \cite{saleh2021probabilistic} is a probabilistic auto-regressive model which consists of two main blocks: MA-Net and the ArTIST model. MA-Net is a recurrent autoencoder that is trained to learn a representation of the dynamical interaction between all agents in the scene. ArTIST is an RNN that takes as input a 4D velocity vector of the current frame for one object as well as the corresponding 256-dimensional interaction representation learned by MA-Net, and outputs a probability distribution for each dimension of the motion velocity for the next frame. As indicated in \cite{saleh2021probabilistic}, the models are trained on the MOT17 training set and the PathTrack \cite{8237302} dataset. We have reused the trained models as well as the tracklet scoring and inpainting code provided by the authors\footnote{available at https://github.com/fatemeh-slh/ArTIST} and reimplemented the object tracking part according to the paper, as this part was not provided. The tracklets are initialized with the bounding boxes detected in the first frame. For any time frame $t$, the score of assigning a detection $\mathbf{o}_{tk}$ to a tracklet $n$ is obtained by evaluating the likelihood of this detection under the distribution estimated by the ArTIST model. The final assignment is computed using the Hungarian algorithm. For any tentatively alive tracklet whose last assignment is prior to $t-1$ with a non-zero gap (implying that there exists a detection absence), the algorithm first performs tracklet inpainting to fill the gap up to $t-1$, then computes the assignment score with the inpainted tracklet. As described in \cite{saleh2021probabilistic}, the inpainting is done with multinominal sampling, and a tracklet rejection scheme (TRS) is applied to select the best inpainted trajectory. In order to eliminate possible inpainting ambiguities, the Hungarian algorithm is run twice, once only for the full sequences without gaps and the second time for the inpainted sequences. The number of candidates for multinominal sampling is set to 50. For the TRS, the IoU threshold used in \cite{saleh2021probabilistic} is $0.5$. In our test scenario, there are less tracklets and the risk of false negative is much greater than that of false positive. So, we decreased the threshold to $0.1$, which provided better results than the original value.

As the proposed DVAE-UMOT algorithm, the VKF algorithm for MOT \cite{8907431} is also based on the VI methodology to combine object position estimation and detection-to-object assignment. However, a basic linear dynamical model is used in VKF instead of the DVAE model in the proposed DVAE-UMOT algorithm. Hence, the VKF MOT algorithm is a combination of VI and Kalman filter update equations.
In \cite{8907431}, the method was proposed in an audiovisual set-up. The observations contain not only the detected bounding box positions, but also appearance features and multichannel audio recordings. For a fair comparison with DVAE-UMOT, we use here the same observations, i.e., we simplified VKF by using only the detected bounding box coordinates. 
\par For both ArTIST and VKF, the tracked sequences are initialized using the detected bounding boxes at the first frame, as what we have done for DVAE-UMOT. For VKF, similarly to DVAE-UMOT, we need to provide initial values for $\mathbf{m}_{tn}$ and $\mathbf{V}_{tn}$. For a fair comparison, we applied the same cascade initialization as the one presented in Section~\ref{subsec:MOT-initialization}, except that for any sub-sequence $\{t_i+1, ..., t_{i+1}\}$ other than the first one, the linear dynamical model is applied to the last frame of the previous sequence to output the initial position vector of the current subsequence. 
The covariance matrices $\mathbf{V}_{tn}$ are initialized with pre-defined values which stabilise the EM algorithm. The observation model covariance matrix are fixed to the same values as for DVAE-UMOT (see Section~\ref{subsec:MOT-initialization}). The covariance matrices of the linear dynamical model (denoted $\boldsymbol{\Lambda}_{tn}$ in \cite{8907431}) are also initialized with the same values.

\subsubsection{Evaluation metrics}

We used the standard MOT metrics \cite{article,ristani2016performance} to evaluate the tracking performance of DVAE-UMOT and compare it to the baselines, namely: multi-object tracking accuracy (MOTA), multi-object tracking precision (MOTP), identity F1 score (IDF1), number of identity switches (IDS), mostly tracked (MT), mostly lost (ML), false positives (FP) and false negatives (FN). The three subsets contain a different number of test sequences, with a different sequence length. Therefore, for IDS, FP and FN, we report both the number of occurrences and the corresponding percentage. Among them, MOTA is considered to be the most representative metric. It is defined by aggregating the frame-wise versions of the metrics FP$_t$, FN$_t$, and ID$_t$ over frames:
\begin{equation}
    \textrm{MOTA} = 1 - \frac{\sum_t(\textrm{FN}_t+\textrm{FP}_t+\textrm{IDS}_t)}{\sum_t \textrm{GT}_t},
    \label{e104}
\end{equation}
where $\textrm{GT}_t$ denotes the number of ground-truth tracks at frame $t$.

Higher MOTA values imply less errors (in terms of FPs, FNs, and IDS), and hence better tracking performance. MOTP defines the averaged overlap between all correctly matched targets and their corresponding ground truth. Higher MOTP implies more accurate position estimations. IDF1 is the ratio of correctly identified detections over the average number of ground-truth and computed detections. IDS reflects the capability of the model to preserve the identity of the tracked objects, especially in case of occlusion and track fragmentation. MT and ML represent how much the trajectory is recovered by the tracking algorithm. A target track is mostly tracked (resp.\ mostly lost) if it is covered by the tracker for at least $80\%$ (resp.\ not more than $20\%$) of its life span.

\subsection{DVAE-UMOT results}

\begin{table*}[!t]
\centering
\caption{Results Obtained by DVAE-UMOT and the Two Baselines on MOT17-3T, for Short ($T = 60$), Medium ($T = 120$), and Long ($T = 300$) Sequences.\label{table:3}\vspace{-4mm}}%
\resizebox{\textwidth}{!}{
 \begin{tabular}{ p{1.3cm} p{2cm} p{1cm} p{1cm} p{0.7cm} p{1cm} p{0.7cm} p{0.7cm} p{0.5cm} p{1cm} p{0.7cm} p{1cm} p{0.7cm} } 
 \toprule
 Dataset & Method & MOTA$\uparrow$ & MOTP$\uparrow$ & IDF1$\uparrow$ & $\#$IDS$\downarrow$ & $\%$IDS$\downarrow$ & MT$\uparrow$  & ML$\downarrow$ & $\#$FP$\downarrow$ & $\%$FP$\downarrow$ & $\#$FN$\downarrow$ & $\%$FN$\downarrow$\\
 \midrule
\multirow{3}*{Short}& ArTIST & 63.7 & \textbf{84.1} & 48.7 & 86371 & 28.0 & \textbf{4684} & \textbf{0} & \textbf{9962} & \textbf{3.2} & \textbf{15525} & \textbf{5.0} \\
 ~ & VKF & 56.0 & 82.7 & 77.3 & 5660 & 1.8 & 3742 & 761 & 64945 & 21.1 & 64945 & 21.1\\
 ~ & DVAE-UMOT & \textbf{79.1} & 81.3 & \textbf{88.4} & \textbf{4966} & \textbf{1.6} & 4370 & 50 & 29808 & 9.7 & 29808 & 9.7\\
 \midrule
 \multirow{3}*{Medium}&ArTIST & 61.0 & \textbf{84.2} & 43.9 & 102978 & 24.6 & \textbf{2943} & \textbf{0} & \textbf{25388} & \textbf{6.1} & \textbf{34812} & \textbf{8.3} \\
 ~& VKF & 57.5 & 83.3 & 77.6 & 7657 & 1.8 & 2563 & 487 & 85053 & 20.3 & 85053 & 20.3 \\
 ~& DVAE-UMOT & \textbf{78.6} & 82.2 & \textbf{88.0} & \textbf{6107} & \textbf{1.5} & 2907 & 120 & 41747 & 9.9 & 41747 & 9.9 \\
 \midrule
 \multirow{3}*{Long}&ArTIST & 53.5 & 84.5 & 40.7 & 205263 & 20.1 & 2513 & \textbf{4} & 135401 & 13.2 & 135401 & 13.2 \\
 ~& VKF & 74.4 & \textbf{86.2} & 84.4 & 30069 & 2.9 & 2756 & 100 & 116160 & 11.4 & 116160 & 11.4 \\
 ~& DVAE-UMOT & \textbf{83.2} & 82.4 & \textbf{90.0} & \textbf{23081} & \textbf{2.3} & \textbf{2890} & 12 & \textbf{74550} & \textbf{7.3} & \textbf{74550} & \textbf{7.3} \\
 \bottomrule
 \end{tabular}}
\end{table*}

\subsubsection{MOT scores}
We now present and discuss the tracking results obtained with the proposed DVAE-UMOT algorithm and compare them with those obtained with the two baselines. 
In these experiments, the value of the observation variance ratio $r_{\boldsymbol{\Phi}}$ is set to 0.04 and no fine-tuning is applied to SRNN in the E-Z step. Ablation study on these factors is presented in Section~\ref{subsec:ablation-study}.

The values of the MOT metrics obtained on short, medium and long sequence subsets ($T=60$, $120$, and $300$ frames, respectively) are shown in Table~\ref{table:3}. We see that the proposed DVAE-UMOT algorithm obtains the best MOTA scores for the three subsets (i.e., for the three different sequence length values). This is remarkable given that ArTIST was trained on the MOT17 training dataset, whereas DVAE-UMOT never saw the ground-truth sequences before the test. Furthermore, we notice that both VKF and DVAE-UMOT have much less IDS and much higher IDF1 scores than ArTIST, which implies that the observation-to-object assignment based on the VI method is more efficient than direct estimation of the position likelihood distribution to preserve the correct object identity during tracking. Besides, the DVAE-UMOT model also has better scores than the VKF model for these two metrics, which implies that the DVAE-based dynamical model performs better on identity preservation than the linear dynamical model of VKF. For the 60- and 120-frame sequences, the ArTIST model has lower FP and FN percentages and higher MOTP scores (though the MOTP scores of all three algorithms are quite close for every value of $T$). This is reasonable because, again, ArTIST was trained on the same dataset using the ground-truth sequences while our model is unsupervised. Overall, the adverse effect caused by frequent identity switches is much greater than the positive effect of lower FP and FN for the ArTIST model. That explains why DVAE-UMOT has much better MOTA scores than ArTIST. However, for the long (300-frame) sequences, DVAE-UMOT obtains an overall much better performance than the ArTIST model, since it obtains here the best scores for 6 metrics out of 8, including FP and FN. This shows that DVAE-UMOT is particularly good at tracking objects on the long term (we recall that $T=300$ represents 10s of video at 30fps). 

Besides, DVAE-UMOT also globally exhibits notably better performance than VKF on all of the three datasets. This clearly indicates that the modeling of the objects dynamics with a DVAE model outperforms the use of a simple linear-Gaussian dynamical model and can greatly improve the tracking performance. We can also notice that the VKF algorithm globally performs much better on 300-frame sequences than on 60- and 120-frame sequences. One possible explanation for this phenomenon is that the dynamical patterns of long sequences are simpler than those of short and medium sequences. This can be verified by looking at the histograms of the tracked objects average velocity in \figurename~\ref{fig:figure4}. The average velocity in long sequences is much lower than in short and medium sequences. In this case, the linear dynamical model can perform quite well (although not as well as the DVAE).

\begin{figure*}[!t]
    \centering
    \subfloat[Example 1: Long-term detection absence.]{\includegraphics[width=\linewidth]{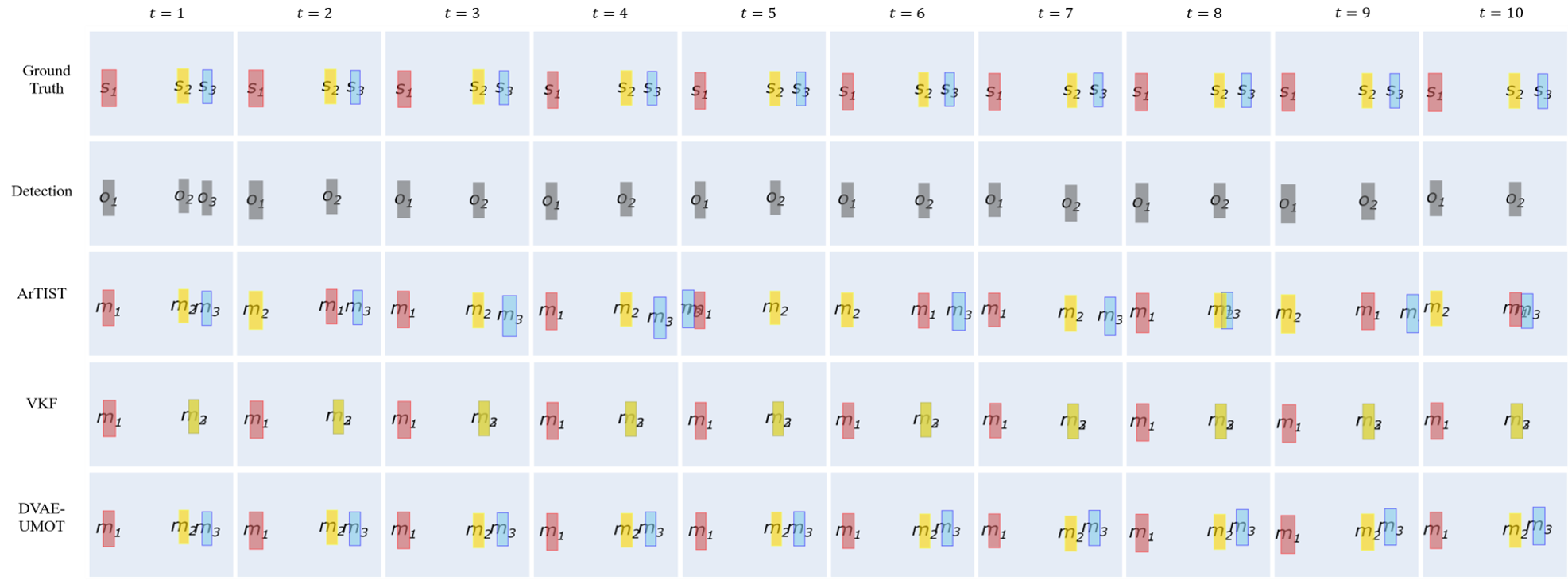}}\\
    \subfloat[Example 2: Crossing objects.]{\includegraphics[width=\linewidth]{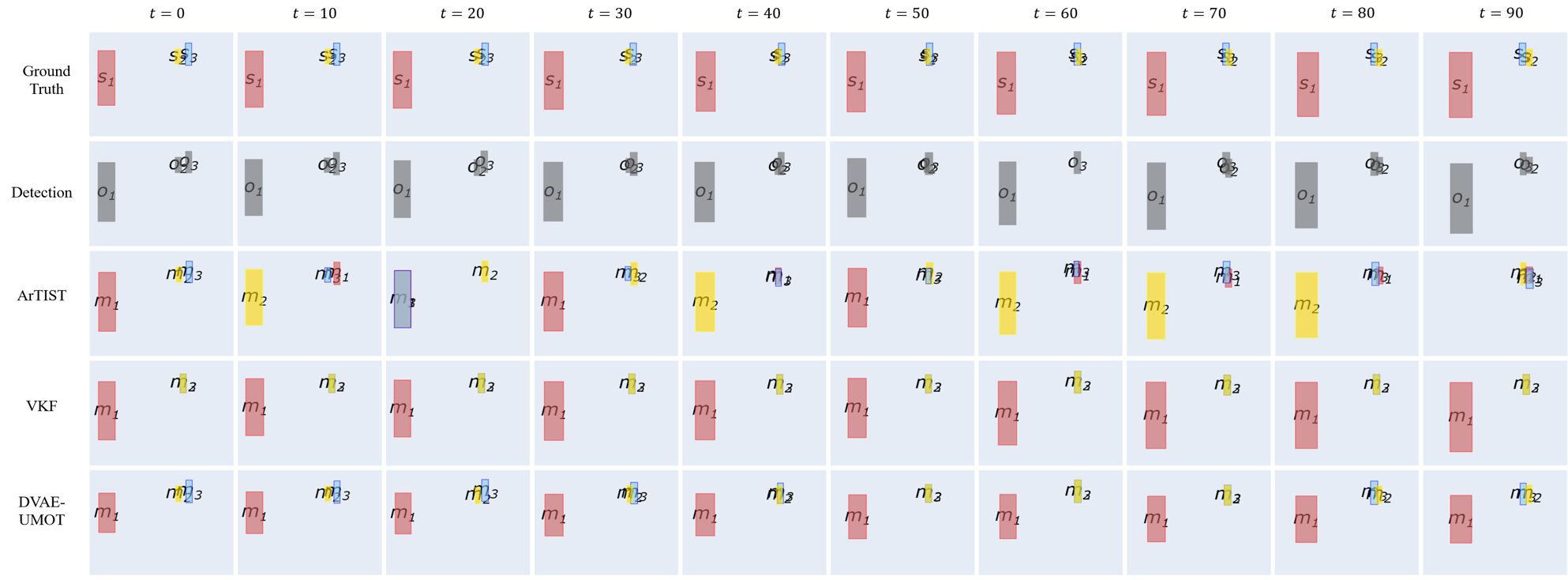}}\\
    \subfloat[Example 3: Crossing objects with frequent detection absence.]{\includegraphics[width=\linewidth]{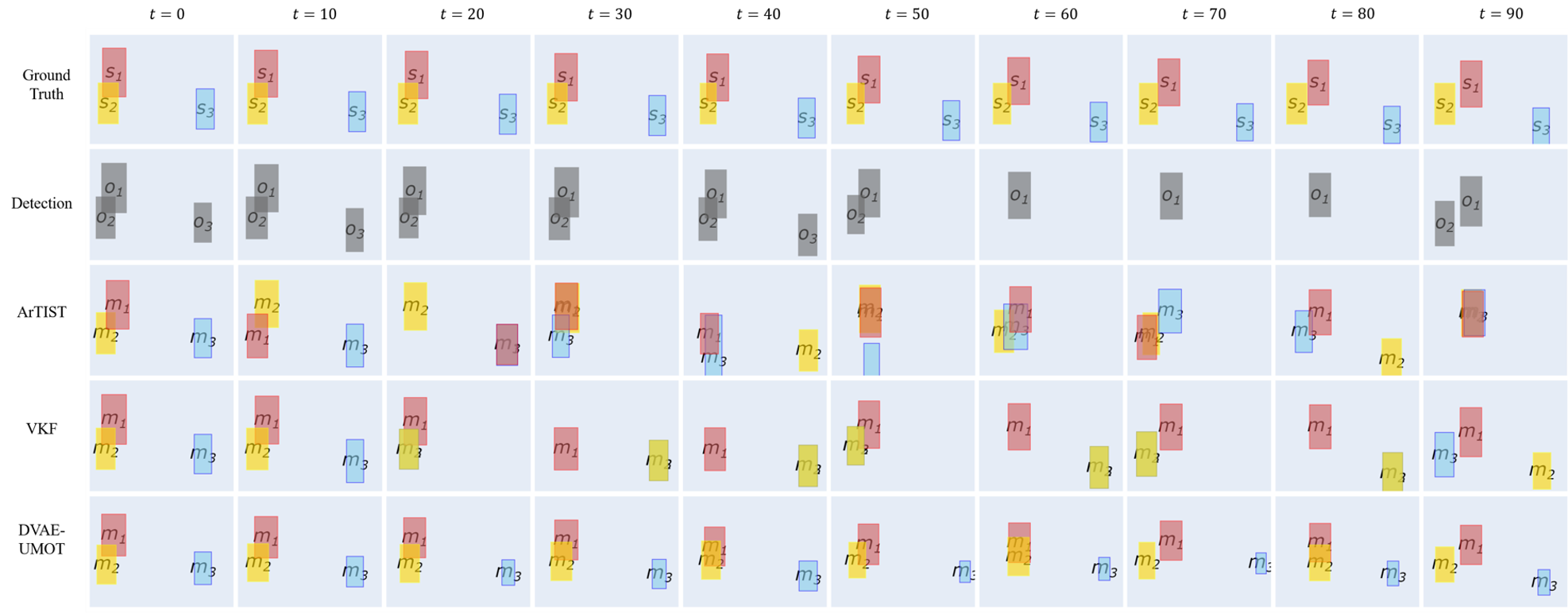}}\\
    \caption{Three examples of tracking result obtained with the proposed DVAE-UMOT algorithm and the two baselines. For clarity of presentation, the simplified notations $s_1$, $o_1$, and $m_1$ denote the ground truth object position, the observation, and the estimated position, respectively (for Object 1, and the same for the two other objects). Best seen in color.}
    \label{fig:figure-examples-tracking-result}
\end{figure*}

\subsubsection{Examples of tracking results}
To illustrate the behavior of each of the three models, we present three graphical examples of tracking results. In the first example plotted in Fig.~\ref{fig:figure-examples-tracking-result} (top), the detection for Object 3 ($o_3$ in the figure) is absent from $t=2$ and reappears after $t=20$. But we limit the plot to $t=10$ for a better visualization. This is a case of long-term detection absence. An immediate identity switch occurs at $t=2$ for the ArTIST model. Then, the track obtained by ArTIST is no longer stable. We speculate the reason for the frequent identity switches made by ArTIST is that the estimated distributions do not correspond well to the true sequential position distributions, which is possibly due to the way these distributions are discretized. Besides the identity switches, the bounding boxes generated by ArTIST at $t=5$, $8$, and $10$ are not accurate. This causes a decrease of the tracking performance. For the VKF model, the estimated bounding boxes for Objects 2 and 3 ($m_2$ and $m_3$ in the figure) overlap each other. This means that the two observations are both assigned to the same object, which is Object 2. From \eqref{eq:posterior-W-eta} and \eqref{eq:posterior-W-beta}, we know that the value of the assignment variable depends on the posterior mean and variance vectors $\mathbf{m}_{tn}$ and $\mathbf{V}_{tn}$, which themselves depend on the dynamical model. With a linear dynamical model, VKF is not able to correctly predict distinct $m_2$ and $m_3$ trajectories. Due to the very good dynamical modeling capacity of the DVAE, DVAE-UMOT still kept tracking despite of the long-term detection absence and generated reasonable bounding boxes for $m_3$, which correspond well to the ground-truth bounding boxes of Object 3 ($s_3$ in the figure).

The second example plotted in Fig.~\ref{fig:figure-examples-tracking-result} (middle) illustrates the case where two persons cross each other. This is one of the most complicated situations that may cause an identity switch and even lead to tracking loss. Considering the limited space for the figure, we display the bounding boxes every ten frames to view the whole process of crossing. For $t=60$, when the ground truth bouding boxes of Objects 2 and 3 ($s_2$ and $s_3$ in the figure) strongly overlap, Detection $o_2$ disappears. Again, ArTIST exhibits frequent identity switches. Besides, at $t=20$, the estimated bounding box $m_1$ is totally overlapped with that of $m_3$. And at $t=90$, the estimated bounding boxes for all of the three objects are getting very close to each other. This indicates that the identity switches can cause unreasonable trajectories estimation. For VKF, the observations for both Object 2 and 3 are assigned to the same target $s_3$ all along the sequence, due to $s_2$ and $s_3$ being close to each other, so that the estimated bounding boxes $m_2$ and $m_3$ overlap completely. In contrast, DVAE-UMOT displays a consistent tracking of the three objects. For $t=50$, $60$, and $70$, the estimated bounding boxes $m_2$ and $m_3$ overlap due to the ground truth bounding boxes $s_2$ and $s_3$ strongly overlap each other. However, the tracking is correctly resumed at $t=80$, with no identity switch (i.e., the crossing of Objects 2 and 3 is correctly captured by the model).

\par The third example displayed in Fig.~\ref{fig:figure-examples-tracking-result} (bottom) is another more complicated situation with two objects very close to each other and frequent detection absence. At $t=20$ when observation $o_3$ disappears, both ArTIST and VKF lose one of the tracks, whereas DVAE-UMOT keeps a reasonable tracking of the three tracks. From $t=60$ to $80$, both $o_2$ and $o_3$ are absent. The tracks inpainted by ArTIST are not consistent anymore and VKF still misses one track. However, even in this difficult scenario, DVAE-UMOT keeps on providing three reasonable trajectories.


\subsection{Ablation study}
\label{subsec:ablation-study}

\subsubsection{Influence of $r_{\boldsymbol{\Phi}}$}

\begin{table*}[!ht]
\centering
\caption{Results Obtained by DVAE-UMOT on MOT17-3T (Short Sequences Subset) for Different Values of $r_{\boldsymbol{\Phi}}$. The Values on the Left (resp. Right) Side of the Slashes are Obtained Without (resp. With) the Fine-tuning of SRNN in the E-Z Step.\vspace{-3mm}}
\label{table:4}
\resizebox{\textwidth}{!}{
 \begin{tabular}[c]{ p{0.02\textwidth} p{0.065\textwidth} p{0.065\textwidth} p{0.065\textwidth} p{0.072\textwidth} p{0.049\textwidth} p{0.072\textwidth} p{0.065\textwidth} p{0.095\textwidth} p{0.065\textwidth} p{0.095\textwidth} p{0.065\textwidth} } 
 \toprule
 $r_{\boldsymbol{\Phi}}$ & MOTA$\uparrow$ & MOTP$\uparrow$ & IDF1$\uparrow$ & $\#$IDs$\downarrow$ & $\%$IDs$\downarrow$ & MT$\uparrow$  & ML$\downarrow$ & $\#$FP$\downarrow$ & $\%$FP$\downarrow$ & $\#$FN$\downarrow$ & $\%$FN$\downarrow$\\
 \midrule
 0.01 & 35.9/32.8 & \textbf{84.5}/\textbf{84.8} & 66.6/65.5 & \textbf{4914}/3216 & \textbf{1.6}/1.0 & 2946/2714 & 916/913 & 96438/102062 & 31.3/33.1 & 96438/102062 & 31.3/33.1 \\
 0.02 & 65.5/61.8 & 84.2/84.7 & 81.3/79.8 & 5319/3073 & 1.7/1.0 & 3932/3652 & 407/379 & 50596/57291 & 16.4/18.6 & 50596/57291 & 16.4/18.6 \\
 0.03 & 74.9/70.0 & 83.1/84.3 & 86.1/84.4 & 5088/\textbf{2853} & 1.7/\textbf{0.9} & 4232/3931 & 158/160 & 36165/43777 & 11.7/14.2 & 36165/43777 & 11.7/14.2 \\
 0.04 & \textbf{79.1}/75.1 & 81.3/83.5 & \textbf{88.4}/86.7 & 4966/2862 & 1.6/0.9 & \textbf{4370}/4067 & 50/64 & \textbf{29808}/36990 & \textbf{9.7}/11.9 & \textbf{29808}/36990 & \textbf{9.7}/11.9 \\
 0.05 & 76.4/\textbf{75.6} & 79.2/82.6 & 87.1/\textbf{87.1} & 4982/2919 & 1.6/0.9 & 4268/\textbf{4066} & \textbf{42}/\textbf{53} & 33924/\textbf{36088} & 11.0/\textbf{11.7} & 33924/\textbf{36088} & 11.0/\textbf{11.7} \\
 0.06 & 69.2/70.1 & 76.9/82.0 & 83.5/84.4 & 5297/3005 & 1.7/1.0 & 3978/3845 & 73/137 & 44793/44598 & 14.5/14.5 & 44793/44598 & 14.5/14.5 \\
 0.07 & 59.8/66.8 & 74.8/80.3 & 78.9/82.9 & 5146/3000 & 1.7/1.0 & 3688/3775 & 188/285 & 59348/49646 & 19.2/16.1 & 59348/49646 & 19.2/16.1 \\
 0.08 & 48.5/60.6 & 73.1/79.4 & 73.3/79.9 & 5097/3119 & 1.7/1.0 & 3303/3637 & 337/432 & 76865/59220 & 24.9/19.2 & 76865/59220 & 24.9/19.2 \\
 \bottomrule
\end{tabular}}
\end{table*}

We start our ablation study by analyzing the influence of $r_{\boldsymbol{\Phi}}$. 
Table~\ref{table:4} reports the MOT scores obtained with DVAE-UMOT as a function of $r_{\boldsymbol{\Phi}}$. These experiments are conducted on the subset of short sequences. We report the results for both with and without fine-tuning SRNN in the E-Z step. Apart from the value of $r_{\boldsymbol{\Phi}}$ and the fine-tuning option, all other conditions are exactly the same across experiments.

Table~\ref{table:4} shows that, whether fine-tuning SRNN in the E-Z step or not, the MOT scores first globally increase with $r_{\boldsymbol{\Phi}}$,\footnote{Except for the MOTP score, which continually decreases with the increase of $r_{\boldsymbol{\Phi}}$. This can be explained as follows. MOTP measures the precision of the position estimation for the matched bounding boxes. The estimated position $\mathbf{m}_{tn}$ in \eqref{eq:posterior-s-mean} is a weighted combination of the observation and the DVAE prediction. When $\boldsymbol{\Phi}_{tk}$ increases, the contribution of the observation decreases and $\mathbf{m}_{tn}$ is closer to the DVAE prediction. Since the error of the DVAE prediction may accumulate over time, this finally decreases the position estimation accuracy.} reach their optimal values for $r_{\boldsymbol{\Phi}} = 0.04$ or $0.05$ (for most metrics), and then decrease for greater $r_{\boldsymbol{\Phi}}$ values. 
For confirmation, we have also computed the averaged empirical ratio $\hat{r}_{\boldsymbol{\Phi}}$ of the detected bounding boxes (with the SDP detector), which is calculated as $\frac{1}{4T}\sum_{t=1}^T\frac{1}{K_t}\sum_{k=1}^{K_t}(\frac{|s^L_{tk} - o^L_{tk}|}{o^R_{tk} - o^L_{tk}} + \frac{|s^T_{tk} - o^T_{tk}|}{o^T_{tk} - o^B_{tk}} + \frac{|s^R_{tk} - o^R_{tk}|}{o^R_{tk} - o^L_{tk}} + \frac{|s^B_{tk} - o^B_{tk}|}{o^T_{tk} - o^B_{tk}})$.\footnote{Note that here $s_{tk}$ denotes the position of the target matched with the observation $o_{tk}$ at time frame $t$. We omit the target positions that are not matched with any observation.} This value equals to $0.053$, $0.053$ and $0.047$ respectively for the short, medium and long sequence dataset. These values,  which are close to each other because we used the same detector, correspond well to the $r_{\boldsymbol{\Phi}}$ value for the best performing model in Table~\ref{table:4}. We can conclude that the model has better performance if the value of $r_{\boldsymbol{\Phi}}$ corresponds (empirically) to the detector performance.

\par
\begin{figure}[!t]
    \centering
    \includegraphics[width=\linewidth]{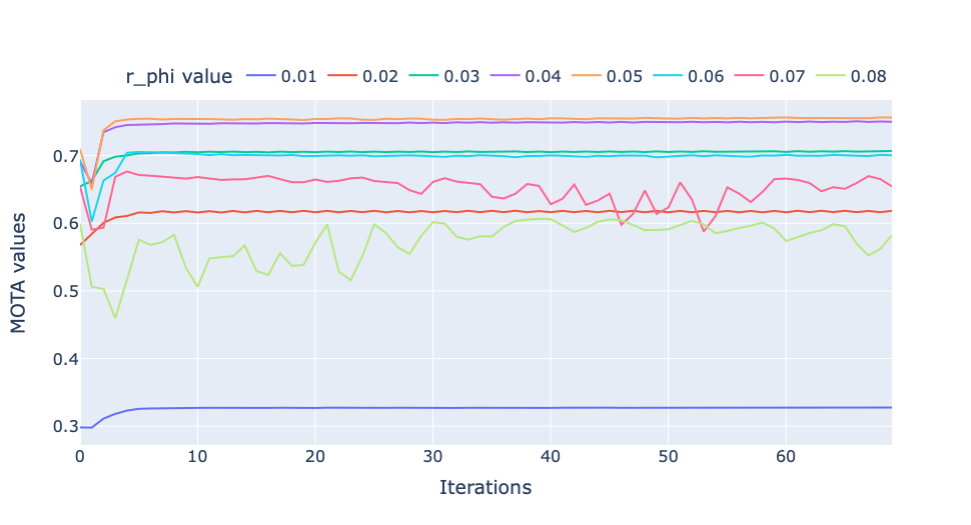}
    \caption{MOTA score obtained by DVAE-UMOT as a function of the number of iterations, for different values of $r_{\boldsymbol{\Phi}}$.}
    \label{fig:figure8}
\end{figure}

 Besides, we have also observed that the value of $r_{\boldsymbol{\Phi}}$ has an impact on the convergence of the DVAE-UMOT algorithm. Fig.~\ref{fig:figure8} displays the MOTA score as a function of the number of DVAE-UMOT iterations (here with the fine-tuning of the DVAE model). It appears clearly that for too high values of $r_{\boldsymbol{\Phi}}$, the model exhibits a lower and more hectic performance than for the optimal value. 

\subsubsection{Influence of  fine-tuning}\label{sec:ablation-causality-finetuning}
\par As mentioned in Section~\ref{sssec:E-Z-step}, SRNN can either be fine-tuned or not in the DVAE-UMOT model. 
Table~\ref{table:5} shows the MOT scores obtained by DVAE-UMOT on the three test subsets with and without the fine-tuning of SRNN in the E-Z step.

\begin{table*}[!ht]
\centering
\caption{Results Obtained by DVAE-UMOT With and Without the Fine-tuning of SRNN. The Results are Reported for the Short, Medium and Long Sequence Test Subsets ($T=60$, $120$, and $300$ Frames, Respectively).\vspace{-3mm}}
\label{table:5}
\resizebox{\textwidth}{!}{
 \begin{tabular}{ p{1.2cm} p{1.8cm} p{1cm} p{1cm} p{0.7cm} p{0.8cm} p{0.7cm} p{0.7cm} p{0.5cm} p{0.8cm} p{0.6cm} p{0.8cm} p{0.6cm} } 
 \toprule
 Dataset & Fine-tuning & MOTA$\uparrow$ & MOTP$\uparrow$ & IDF1$\uparrow$ & $\#$IDs$\downarrow$ & $\%$IDs$\downarrow$ & MT$\uparrow$  & ML$\downarrow$ & $\#$FP$\downarrow$ & $\%$FP$\downarrow$ & $\#$FN$\downarrow$ & $\%$FN$\downarrow$\\
 \midrule
 \multirow{2}*{Short}& Fine-tune & 75.1 & \textbf{83.5} & 86.7 & \textbf{2862} & \textbf{0.9} & 4067 & 64 & 36990 & 11.9 & 36990 & 11.9\\
 ~ & No fine-tune & \textbf{79.1} & 81.3 & \textbf{88.4} & 4966 & 1.6 & \textbf{4370} & \textbf{50} & \textbf{29808} & \textbf{9.7} & \textbf{29808} & \textbf{9.7}\\
 \midrule
 \multirow{2}*{Medium}& Fine-tune & 73.1 & \textbf{84.0} & 85.9 & \textbf{3044} & \textbf{0.7} & 2705 & 136 & 54604 & 13.1 & 54604 & 13.1\\
 ~ & No fine-tune & \textbf{78.6} & 82.2 & \textbf{88.0} & 6107 & 1.5 & \textbf{2907} & \textbf{120} & \textbf{41747} & \textbf{9.9} & \textbf{41747} & \textbf{9.9}\\
  \midrule
 \multirow{2}*{Long}& Fine-tune & 65.6 & \textbf{84.9} & 81.6 & \textbf{8670} & \textbf{0.8} & 2286 & 67 & 171515 & 13.8 & 171515 & 13.8\\
 ~ & No fine-tune & \textbf{83.2} & 82.4 & \textbf{90.0} & 23081 & 2.3 & \textbf{2890} & \textbf{12} & \textbf{74550} & \textbf{7.3} & \textbf{74550} & \textbf{7.3}\\
 \bottomrule
 \end{tabular}}

\end{table*}

\par We observe that for all of the three datasets, not fine-tuning the DVAE model obtains the best overall performance (as measured by MOTA in particular). Though fine-tuning the DVAE model can indeed increase the MOTP score and decrease the number of identity switches, it does not improve the overall tracking performance. Indeed, fine-tuning increases the FP and FN numbers/proportions, and thus decreases the MOTA scores. Especially on the long sequence dataset, the MOTA score drops from $83.2$ to $65.6$. The possible reason is that fine-tuning could make the model more sensible to detection noise, and lead to a generative model with worse performance.

\section{Conclusion and Future Work}
In this paper, we have proposed a deep LVGM and corresponding algorithm for MOT, called DVAE-UMOT. 
This method is based on VI at two levels. The first level is the pre-training of a DVAE model (in our experiments SRNN) on a synthetic single-object dataset; The DVAE is used within DVAE-UMOT as a deep probabilistic model of object dynamics. The second level is the estimation of the parameters of the overall DVAE-UMOT model, which is done on every test multi-object sequence to be processed.  
Observation-to-track association is done by estimating an assignment variable distribution while each object position distribution is estimated by combining the detection and the DVAE prediction. Our model is unsupervised, i.e., it does not use any manual annotation of ground-truth object position sequence for training, in fact it does not use any labeled dataset at all. 
\par We evaluated the performance of the proposed algorithm by conducting experiments on a multi-object dataset derived from the MOT17 training set. Experimental results show that DVAE-UMOT obtains MOTA scores that are notably higher than those obtained by two state-of-the-art baselines. In particular, it largely reduces the identity switches compared to ArTIST and obtain consistently better results than VKF for almost all metrics. In case of detection absence and frequent occlusions, DVAE-UMOT can keep good tracking performance and generate reasonable bounding boxes to fill in the detection gaps. Compared to the linear motion model used in VKF (and in many other Kalman filter-based methods), the DVAE-based model is able to model more complex object dynamics and thus proves more accurate and robust for MOT in the VI framework. Finally, we conducted ablation studies to better understand the influence of several factors/settings on the DVAE-UMOT performance. 
\par Of course, the quality of the visual information is essential to design a good MOT model. We demonstrated in this work that under the tracking-by-detection paradigm, modeling the motion information with a powerful non-linear DVAE model can lead to achieve quite good tracking performance. In the present study, we did not consider the object appearance information, and did not consider the track birth/death process as well, because we wanted the study to focus on the potential of the DVAE for modeling the object dynamics in a MOT framework. The design of a complete MOT system based on a DVAE should consider those two aspects, which is planned in our future works. 



\bibliographystyle{IEEEtran}
\bibliography{IEEEabrv,ref}

\cleardoublepage
\appendices
\begin{figure*}[t]
\begin{minipage}{\textwidth}
\begin{center}
{\huge
Supplementary material of\\ Unsupervised multiple-object tracking\vspace{1mm}\\ with a dynamical variational autoencoder
}\vspace{3mm}\\
{\large Xiaoyu~Lin, Laurent~Girin, Xavier~Alameda-Pineda, \textit{IEEE Senior Member}}
\end{center}
\end{minipage}
\end{figure*}
\renewcommand{\thesection}{\Alph{section}}
\setcounter{section}{0}

\section{DVAE-UMOT algorithm calculation details}\label{appe:VEM-calculations}
\subsection{E-S Step}
\label{appe:E-S-step}
Here we detail the calculation of the posterior distribution $q_{\phi_{\mathbf{s}}}(\mathbf{s}|\mathbf{o})$. Using \eqref{eq:MOT-obs-factorized}, the first expectation term in \eqref{eq:optimal_qs} can be developed as:
\begin{align*}
    & \mathbb{E}_{q_{\phi_{\mathbf{w}}}(\mathbf{w}|\mathbf{o})}\big[\log p_{\theta_{\mathbf{o}}}(\mathbf{o} | \mathbf{w}, \mathbf{s})\big]\nonumber
    \\ & \qquad = \mathbb{E}_{q_{\phi_{\mathbf{w}}}(\mathbf{w}|\mathbf{o})}\bigg[\sum_{t=1}^T \sum_{k=1}^{K_t} \log p_{\theta_{\mathbf{o}}}(\mathbf{o}_{tk} | w_{tk}, \mathbf{s}_{t, 1:N})\bigg] \\
    & \qquad = \sum_{t=1}^T \sum_{k=1}^{K_t} \mathbb{E}_{q_{\phi_{\mathbf{w}}}(w_{tk}|\mathbf{o}_{tk})} \big[\log p_{\theta_{\mathbf{o}}}(\mathbf{o}_{tk} | w_{tk}, \mathbf{s}_{t, 1:N})\big].
\end{align*}
Since for any pair $(t, k)$, the assignment variable $w_{tk}$ follows a discrete posterior distribution, we can denote its values by
\[
    q_{\phi_{\mathbf{w}}}(w_{tk} = n|\mathbf{o}_{tk}) = \eta_{tkn},\label{eq:appe-e32}
\]
which will be calculated later in the E-W Step. With this notation, we have:
\begin{align*}
    & \mathbb{E}_{q_{\phi_{\mathbf{w}}}(\mathbf{w}|\mathbf{o})}\big[\log p_{\theta_{\mathbf{o}}}(\mathbf{o} | \mathbf{w}, \mathbf{s})\big]\nonumber
    \\ & \qquad \qquad = \sum_{t=1}^T \sum_{k=1}^{K_t} \sum_{n=1}^N \eta_{tkn} \log p_{\theta_{\mathbf{o}}}(\mathbf{o}_{tk} | w_{tk}=n, \mathbf{s}_{tn}).
    \label{eq:appe-e33} 
\end{align*}
The second expectation in~\eqref{eq:optimal_qs} cannot be computed analytically as a distribution on $\mathbf{s}$ because of the non-linearity in the decoder and in the encoder. In order to avoid a tedious sampling procedure and obtain a computationally efficient solution, we further approximate this term by assuming  $q_{\phi_{\mathbf{z}}}(\mathbf{z} | \mathbf{s})\approx q_{\phi_{\mathbf{z}}}(\mathbf{z} | \mathbf{s}=\mathbf{m}^{(i-1)})$, where $\mathbf{m}^{(i-1)}$ is the mean value of the posterior distribution of $\mathbf{s}$ estimated at the previous iteration. By using this approximation, the term $\mathbb{E}_{q_{\phi_{\mathbf{z}}}(\mathbf{z} | \mathbf{s})}\big[\log q_{\phi_{\mathbf{z}}}(\mathbf{z} | \mathbf{s})\big]$ is now considered as a constant. 

In addition, we observe that the second term of~\eqref{eq:optimal_qs} can be rewritten as:
\begin{align*}
    & \mathbb{E}_{q_{\phi_{\mathbf{z}}}(\mathbf{z} | \mathbf{s})}\big[\log p_{\theta_{\mathbf{s}\mathbf{z}}}(\mathbf{s}, \mathbf{z})\big]\nonumber
    \\ & \qquad \qquad \qquad= \sum_{n=1}^N \mathbb{E}_{q_{\phi_{\mathbf{z}}}(\mathbf{z}_{:, n} | \mathbf{m}_{:, n}^{(i-1)})}\big[\log p_{\theta_{\mathbf{s}\mathbf{z}}}(\mathbf{s}_{:, n}, \mathbf{z}_{:, n})\big],
\end{align*}
since both the DVAE joint distribution and posterior distribution factorise over the objects, as formalized  in \eqref{eq:MOT-DVAE-factorized-over-n} and \eqref{eq:qz_factorises}.
As a consequence, the posterior distribution of $\mathbf{s}$ factorises over the tracked object:
\[
    q_{\phi_{\mathbf{s}}}(\mathbf{s}|\mathbf{o}) 
    = \prod_{n=1}^N q_{\phi_{\mathbf{s}}}(\mathbf{s}_{:, n}|\mathbf{o}), \label{eq:appe-e37}
\]
and therefore:
\begin{multline}
    q_{\phi_{\mathbf{s}}}(\mathbf{s}_{:, n}|\mathbf{o}) \propto \exp \Big(\mathbb{E}_{q_{\phi_{\mathbf{z}}}(\mathbf{z}_{:, n} | \mathbf{m}_{:, n}^{(i-1)})}\big[\log p_{\theta_{\mathbf{s}\mathbf{z}}}(\mathbf{s}_{:, n}, \mathbf{z}_{:, n})\big]\Big)\\ \prod_{t=1}^T \prod_{k=1}^{K_t} \exp \big(\eta_{tkn} \log p_{\theta_{\mathbf{o}}}(\mathbf{o}_{tk}|w_{tk}=n, \mathbf{s}_{tn})\big).\nonumber
\end{multline}

In the above equation, the expectation term cannot be calculated in closed form. As usually done in the DVAE methodology, it is thus replaced by a Monte Carlo estimate using sampled sequences drawn from the DVAE inference model. Let us denote by $\mathbf{z}_{:, n}^{(i)} \sim q_{\phi_{\mathbf{z}}}(\mathbf{z}_{:, n} | \mathbf{m}_{:, n}^{(i-1)})$ such a sampled sequence. In the present work, we use single point estimate, thus obtaining:
\begin{align*}
    & q_{\phi_{\mathbf{s}}}(\mathbf{s}_{:, n}|\mathbf{o}) \nonumber
    \\ & \propto p_{\theta_{\mathbf{s}\mathbf{z}}}(\mathbf{s}_{:, n}, \mathbf{z}_{:, n}^{(i)})\prod_{t=1}^T \prod_{k=1}^{K_t} \exp \big(\eta_{tkn} \log p_{\theta_{\mathbf{o}}}(\mathbf{o}_{tk}|w_{tk}=n, \mathbf{s}_{tn})\big) \\
    & \propto \prod_{t=1}^T\Big(p_{\theta_{\mathbf{s}}}(\mathbf{s}_{tn} | \mathbf{s}_{1:t-1, n}, \mathbf{z}_{1:t, n}^{(i)})p_{\theta_{\mathbf{z}}}(\mathbf{z}_{tn}^{(i)} | \mathbf{s}_{1:t-1, n}, \mathbf{z}_{1:t-1, n}^{(i)})\nonumber \\ 
    & \qquad \qquad \quad \prod_{k=1}^{K_t}\exp \big(\eta_{tkn} \log p_{\theta_{\mathbf{o}}}(\mathbf{o}_{tk}|w_{tk}=n, \mathbf{s}_{tn})\big)\Big).
\end{align*}

We observe that the $t$-th element of the previous factorisation is a distribution over $\mathbf{s}_{tn}$ conditioned by $\mathbf{s}_{1:t-1,n}$. As for $q_{\phi_{\mathbf{z}}}(\mathbf{z}_{:, n} | \mathbf{s}_{:, n})$, the dependency with $\mathbf{s}_{1:t-1,n}$ is non-linear and therefore would impede to obtain a computationally efficient closed-form solution. In the same attempt of avoiding costly sampling strategies, we approximate the previous expression replacing $\mathbf{s}_{1:t-1,n}$ with $\mathbf{s}_{1:t-1,n}^{(i)}$, obtaining:
\[
    q_{\phi_{\mathbf{s}}}(\mathbf{s}_{:, n}|\mathbf{o}) \approx  \prod_{t=1}^T q_{\phi_{\mathbf{s}}}(\mathbf{s}_{tn} | \mathbf{s}_{1:t-1, n}^{(i)} ,\mathbf{o}),
\]
with
\begin{align*}
    q_{\phi_{\mathbf{s}}}(\mathbf{s}_{tn} | \mathbf{s}_{1:t-1, n}^{(i)} ,\mathbf{o}) & \propto p_{\theta_{\mathbf{s}}}(\mathbf{s}_{tn} | \mathbf{s}_{1:t-1, n}^{(i)}, \mathbf{z}_{1:t, n}^{(i)}) \nonumber \\ & \prod_{k=1}^{K_t}\exp \big(\eta_{tkn} \log p_{\theta_{\mathbf{o}}}(\mathbf{o}_{tk}|w_{tk}=n, \mathbf{s}_{tn})\big),\label{eq:appe-e40}
\end{align*}
since the term $p_{\theta_{\mathbf{z}}}(\mathbf{z}_{tn}^{(i)} | \mathbf{s}_{1:t-1, n}^{(i)}, \mathbf{z}_{1:t-1, n}^{(i)})$ becomes a constant.

Another interesting consequence of sampling $\mathbf{s}_{1:t-1,n}$ is that the dependency with the future observations of $q_{\phi_{\mathbf{s}}}(\mathbf{s}_{tn} | \mathbf{s}_{1:t-1, n}^{(i)} ,\mathbf{o})$ disappears. Indeed, since we are sampling \textit{at every time step}, the future posterior distributions $q_{\phi_{\mathbf{s}}}(\mathbf{s}_{t+k,n} | \mathbf{s}_{1:t+k-1, n}^{(i)} ,\mathbf{o})$ do not depend on $\mathbf{s}_{tn}$, and therefore the posterior distribution of $\mathbf{s}_{tn}$ will not depend on the future observations.



The two distributions in the above equation are Gaussian distributions defined in (\ref{eq:DVAE-generative-distr-s}), and (\ref{eq:MOT-obs-Gaussian}). Therefore, it can be shown that the variational posterior distribution of $\mathbf{s}_{tn}$ is a Gaussian distribution:  $q_{\phi_{\mathbf{s}}}(\mathbf{s}_{tn}| \mathbf{s}_{1:t-1, n}^{(i)}, \mathbf{o}) = \mathcal{N}(\mathbf{s}_{tn} ; \mathbf{m}_{tn}, \mathbf{V}_{tn})$ 
with covariance matrix and mean vector provided in (\ref{eq:posterior-s-cov}) and (\ref{eq:posterior-s-mean}) respectively, and recalled here for completeness:
\[
    \mathbf{V}_{tn} = \Big(\textstyle \sum\limits_{k=1}^{K_t} \eta_{tkn}  \boldsymbol{\Phi}_{tk}^{-1} + \textrm{diag} (\boldsymbol{v}^{(i)}_{\theta_{\mathbf{s}},tn})^{-1}\Big)^{-1},
\]
\[ 
    \mathbf{m}_{tn} = \mathbf{V}_{tn} \Big(\textstyle \sum\limits_{k=1}^{K_t} \eta_{tkn}  \boldsymbol{\Phi}_{tk}^{-1} \mathbf{o}_{tk} + \textrm{diag} (\boldsymbol{v}^{(i)}_{\theta_{\mathbf{s}},tn})^{-1} \boldsymbol{\mu}^{(i)}_{\theta_{\mathbf{s}},tn}\Big),
\]
where $\boldsymbol{v}^{(i)}_{\theta_{\mathbf{s}},tn}$ and $\boldsymbol{\mu}^{(i)}_{\theta_{\mathbf{s}},tn}$ are simplified notations for $\boldsymbol{v}_{\theta_{\mathbf{s}}}(\mathbf{s}_{1:t-1, n}^{(i)}, \mathbf{z}_{1:t, n}^{(i)})$ and $\boldsymbol{\mu}_{\theta_{\mathbf{s}}}(\mathbf{s}_{1:t-1, n}^{(i)}, \mathbf{z}_{1:t, n}^{(i)})$ respectively, denoting the variance and mean vector estimated by the DVAE for object $n$ at time frame $t$.
\subsection{E-Z Step}
\label{appe:E-Z-VLB}
Here we detail the calculation of the ELBO term      \eqref{eq:DVAE-VLB-in-MOT}.
\begin{align*}
    & \mathcal{L}(\theta_{\mathbf{s}}, \theta_{\mathbf{z}}, \phi_{\mathbf{z}}; \mathbf{o})\nonumber\\
    & \quad = \mathbb{E}_{q_{\phi_{\mathbf{s}}}(\mathbf{s}|\mathbf{o})}\Big[\mathbb{E}_{q_{\phi_{\mathbf{z}}}(\mathbf{z} | \mathbf{s})}\big[\log p_{\theta_{\mathbf{s}\mathbf{z}}}(\mathbf{s}, \mathbf{z}) - \log q_{\phi_{\mathbf{z}}}(\mathbf{z} | \mathbf{s})\big]\Big] \\
    & \quad = \mathbb{E}_{\prod\limits_{n=1}^{N} q_{\phi_{\mathbf{s}}}(\mathbf{s}_{:,n}|\mathbf{o})} \bigg[\mathbb{E}_{\prod\limits_{n=1}^{N} q_{\phi_{\mathbf{z}}}(\mathbf{z}_{:,n} | \mathbf{s}_{:,n})} \Big[\textstyle \sum\limits_{n=1}^N \log p_{\theta_{\mathbf{s}\mathbf{z}}}(\mathbf{s}_{:,n}, \mathbf{z}_{:,n})\Big] \nonumber \\
    & \qquad \qquad - \mathbb{E}_{\prod\limits_{n=1}^{N} q_{\phi_{\mathbf{z}}}(\mathbf{z}_{:,n} | \mathbf{s}_{:,n})} \Big[\textstyle \sum\limits_{n=1}^N \log q_{\phi_{\mathbf{z}}}(\mathbf{z}_{:,n} | \mathbf{s}_{:,n})\Big]\bigg]\\
    & = \sum_{n=1}^N  \mathbb{E}_{ q_{\phi_{\mathbf{s}}}(\mathbf{s}_{:,n}|\mathbf{o})} \Big[ \mathbb{E}_{q_{\phi_{\mathbf{z}}}(\mathbf{z}_{:,n} | \mathbf{s}_{:,n})} \big[\log p_{\theta_{\mathbf{s}\mathbf{z}}}(\mathbf{s}_{:,n}, \mathbf{z}_{:,n})\big] \nonumber\\ 
    & \qquad \qquad \qquad \qquad - \mathbb{E}_{ q_{\phi_{\mathbf{z}}}(\mathbf{z}_{:,n} | \mathbf{s}_{:,n})} \big[\log q_{\phi_{\mathbf{z}}}(\mathbf{z}_{:,n} | \mathbf{s}_{:,n})\big] \Big] \\
    & = \sum_{n=1}^N \mathcal{L}_{n}(\theta_{\mathbf{s}}, \theta_{\mathbf{z}}, \phi_{\mathbf{z}}; \mathbf{o}),
\end{align*}
with
\begin{align*}
    & \mathcal{L}_{n}(\theta_{\mathbf{s}}, \theta_{\mathbf{z}}, \phi_{\mathbf{z}}; \mathbf{o}) \nonumber \\
    & \qquad \qquad =  \mathbb{E}_{ q_{\phi_{\mathbf{s}}}(\mathbf{s}_{:,n}|\mathbf{o})} \Big[ \mathbb{E}_{q_{\phi_{\mathbf{z}}}(\mathbf{z}_{:,n} | \mathbf{s}_{:,n})} \big[\log p_{\theta_{\mathbf{s}\mathbf{z}}}(\mathbf{s}_{:,n}, \mathbf{z}_{:,n})\big] \nonumber \\
    & \qquad \qquad \qquad \quad - \mathbb{E}_{ q_{\phi_{\mathbf{z}}}(\mathbf{z}_{:,n} | \mathbf{s}_{:,n})} \big[\log q_{\phi_{\mathbf{z}}}(\mathbf{z}_{:,n} | \mathbf{s}_{:,n})\big] \Big]. 
\end{align*}

\subsection{E-W Step}
Here we detail the calculation of the posterior distribution $q_{\phi_{\mathbf{w}}}(\mathbf{w}|\mathbf{o})$. Applying the optimal update equation~\eqref{eq:optimal-factorized-posterior-general} to $\mathbf{w}$, we have:
\begin{align*}
    q_{\phi_{\mathbf{w}}}(\mathbf{w} | \mathbf{o})
    & \propto \exp \Big(\mathbb{E}_{q_{\phi_{\mathbf{s}}}(\mathbf{s}|\mathbf{o})q_{\phi_{\mathbf{z}}}(\mathbf{z} | \mathbf{s})}\big[\log p_{\theta}(\mathbf{o}, \mathbf{w}, \mathbf{s}, \mathbf{z})\big]\Big). 
\end{align*}
Using \eqref{eq:MOT-generative-factorized}, we derive:
\[
    q_{\phi_{\mathbf{w}}}(\mathbf{w} | \mathbf{o}) \propto p_{\theta_{\mathbf{w}}}(\mathbf{w}) \exp \Big(\mathbb{E}_{q_{\phi_{\mathbf{s}}}(\mathbf{s}|\mathbf{o})} \big[\log p_{\theta_{\mathbf{o}}}(\mathbf{o} | \mathbf{w}, \mathbf{s})\big]\Big).
\]
Using \eqref{eq:MOT-obs-factorized}, the expectation term can be developed as:\footnote{In fact, the posterior distribution $q_{\phi_{\mathbf{s}}}(\mathbf{s}_{t, :}|\mathbf{o})$ is also conditioned on $\mathbf{s}_{1:t-1, :}$ and $\mathbf{z}_{1:t, :}$. We use this abuse of notation for concision.}
\begin{align*}
    & \mathbb{E}_{q_{\phi_{\mathbf{s}}}(\mathbf{s}|\mathbf{o})} \big[\log p_{\theta_{\mathbf{o}}}(\mathbf{o} | \mathbf{w}, \mathbf{s})\big]\nonumber \\
    & \qquad \qquad= \mathbb{E}_{q_{\phi_{\mathbf{s}}}(\mathbf{s}|\mathbf{o})} \Big[\textstyle \sum\limits_{t=1}^T \sum\limits_{k=1}^{K_t} \log p_{\theta_{\mathbf{o}}}(\mathbf{o}_{tk}| w_{tk}, \mathbf{s}_{t, :})\Big]\\
    & \qquad \qquad= \sum_{t=1}^T \sum_{k=1}^{K_t} \mathbb{E}_{q_{\phi_{\mathbf{s}}}(\mathbf{s}_{t, :}|\mathbf{o})} \big[\log p_{\theta_{\mathbf{o}}}(\mathbf{o}_{tk}| w_{tk}, \mathbf{s}_{t, :})\big].
\end{align*} 
Combining \eqref{eq:MOT-assign-factorized} and the previous result, we have:
\begin{align*}
    & q_{\phi_{\mathbf{w}}}(\mathbf{w} | \mathbf{o})\nonumber\\
    & \propto \prod_{t=1}^T \prod_{k=1}^{K_t} p_{\theta_{\mathbf{w}}}(w_{tk}) \exp \Big(\mathbb{E}_{q_{\phi_{\mathbf{s}}}(\mathbf{s}_{t, :}|\mathbf{o})} \big[\log p_{\theta_{\mathbf{o}}}(\mathbf{o}_{tk}| w_{tk}, \mathbf{s}_{t, :})\big] \Big),
\end{align*}
which we can rewrite
\begin{align*}
    q_{\phi_{\mathbf{w}}}(\mathbf{w} | \mathbf{o}) & \propto \prod_{t=1}^T \prod_{k=1}^{K_t} q_{\phi_{\mathbf{w}}}(w_{tk}|\mathbf{o}),
\end{align*}
with
\begin{align*}
 & q_{\phi_{\mathbf{w}}}(w_{tk}|\mathbf{o})\nonumber \\
 & = p_{\theta_{\mathbf{w}}}(w_{tk}) \exp \Big(\mathbb{E}_{q_{\phi_{\mathbf{s}}}(\mathbf{s}_{t, :}|\mathbf{o})} \big[\log p_{\theta_{\mathbf{o}}}(\mathbf{o}_{tk}| w_{tk}, \mathbf{s}_{t, :})\big] \Big).
\end{align*}
The assignment variable $w_{tk}$ follows a discrete distribution and we denote: 
\begin{align*}
    & \eta_{tkn} = q_{\phi_{\mathbf{w}}}(w_{tk} = n|\mathbf{o}) \\
    &\propto p_{\phi_{\mathbf{w}}}(w_{tk}=n) \exp \Big(\mathbb{E}_{q_{\phi_{\mathbf{s}}}(\mathbf{s}_{tn}|\mathbf{o})} \big[\log p_{\theta_{\mathbf{o}}}(\mathbf{o}_{tk}| w_{tk} = n, \mathbf{s}_{tn}) \big] \Big).
\end{align*}
Using the fact that both  $p_{\theta_{\mathbf{o}}}(\mathbf{o}_{tk}| w_{tk} = n, \mathbf{s}_{tn})$ and $q_{\phi_{\mathbf{s}}}(\mathbf{s}_{tn}|\mathbf{o})$ are  multivariate Gaussian distributions (defined in \eqref{eq:MOT-obs-Gaussian} and \eqref{eq:posterior-s-Gaussian}--\eqref{eq:posterior-s-mean}, respectively), the previous expectation can be calculated in closed form:
\begin{align*}
    & \mathbb{E}_{q_{\phi_{\mathbf{s}}}(\mathbf{s}_{tn}|\mathbf{o})} \big[\log p_{\theta_{\mathbf{o}}}(\mathbf{o}_{tk}| w_{tk} = n, \mathbf{s}_{tn})\big] \nonumber\\
    & = \int_{\mathbf{s}_{tn}} \mathcal{N}(\mathbf{s}_{tn} ; \mathbf{m}_{tn}, \mathbf{V}_{tn}) \log \mathcal{N} (\mathbf{o}_{tk}; \mathbf{s}_{tn}, \boldsymbol{\Phi}_{tk}) d\mathbf{s}_{tn}, \\
    & = -\frac{1}{2} \Big[\log |\boldsymbol{\Phi}_{tk}| + (\mathbf{o}_{tk} - \mathbf{m}_{tn})^T \boldsymbol{\Phi}^{-1}_{tk}(\mathbf{o}_{tk} - \mathbf{m}_{tn}) \nonumber\\ 
    & \qquad \qquad \qquad \qquad \qquad \qquad \quad + \text{Tr}\big(\boldsymbol{\Phi}^{-1}_{tk} \mathbf{V}_{tn}\big)\Big].
\end{align*}
By using \eqref{eq:MOT-assign-uniform}, the previous result, and normalizing to 1, we finally get:
\[
    \eta_{tkn} = \frac{\beta_{tkn}}{\sum_{i=1}^N \beta_{tki}},
\]
where
\[
    \beta_{tkn} = \mathcal{N}(\mathbf{o}_{tk} ; \mathbf{m}_{tn}, \boldsymbol{\Phi}_{tk})\exp \Big(-\frac{1}{2}\text{Tr}\big( \boldsymbol{\Phi}^{-1}_{tk} \mathbf{V}_{tn}\big)\Big).
\]

\subsection{M Step}
Here we detail the calculation of $\boldsymbol{\Phi}_{tk}$. In the ELBO expression \eqref{eq:elbo}, only the first term depends on $\theta_{\mathbf{o}}$:
\begin{align*}
    \mathcal{L}(\theta_{\mathbf{o}};\mathbf{o}) \nonumber
    &= \mathbb{E}_{q_{\phi_{\mathbf{w}}}(\mathbf{w}|\mathbf{o})q_{\phi_{\mathbf{s}}}(\mathbf{s}|\mathbf{o})}\big[\log p_{\theta_{\mathbf{o}}}(\mathbf{o} | \mathbf{w}, \mathbf{s})\big] \\
    & = \sum_{n=1}^N \sum_{t=1}^T \sum_{k=1}^{K_t} \eta_{tkn} \int_{\mathbf{s}_{tn}} \mathcal{N}(\mathbf{s}_{tn} ; \mathbf{m}_{tn}, \mathbf{V}_{tn})\nonumber\\
    & \qquad \qquad \qquad \qquad \qquad \qquad \log \mathcal{N} (\mathbf{o}_{tk}; \mathbf{s}_{tn}, \boldsymbol{\Phi}_{tk}) d\mathbf{s}_{tn}\\
    & = - \frac{1}{2} \sum_{n=1}^N \sum_{t=1}^T \sum_{k=1}^{K_t} \eta_{tkn} \Big[\log |\boldsymbol{\Phi}_{tk}| \nonumber\\
    & + (\mathbf{o}_{tk} - \mathbf{m}_{tn})^T \boldsymbol{\Phi}^{-1}_{tk}(\mathbf{o}_{tk} - \mathbf{m}_{tn}) + \text{Tr}( \boldsymbol{\Phi}^{-1}_{tk} \mathbf{V}_{tn})\Big].
\end{align*}
By computing the derivative of $\mathcal{L}(\theta_{\mathbf{o}};\mathbf{o})$ with respect to $\boldsymbol{\Phi}_{tk}$ and setting it to 0, we find the optimal value of $\boldsymbol{\Phi}_{tk}$ that maximizes the ELBO:
\begin{align*}
     \boldsymbol{\Phi}_{tk} &= \sum_{n=1}^N \eta_{tkn}\Big((\mathbf{o}_{tk} - \mathbf{m}_{tn})(\mathbf{o}_{tk} - \mathbf{m}_{tn})^T 
     + \mathbf{V}_{tn}\Big).
\end{align*}

\section{Cascade initialization of the position vector sequence }\label{appe:cascade-init}
For the initialization of the position vector, we first split the long sequence indexed by $t \in \{1, 2, ..., T\}$ into $J$ smaller sub-sequences indexed by $\{\{1, ..., t_1\}, \{t_1+1, ..., t_2\}, ..., \{t_{J-1}+1, ..., T\}\}$. For the first sub-sequence, the mean vector sequence $\mathbf{m}_{1:t_1,n}$ is initialized as the detected vector at the first frame $\mathbf{o}_{1k}$ repeated for $t_1$ times with a arbitrary order of assignment. Thus, there are as many tracked sources as initial detections, i.e., this implicitly sets $N=K_1$. The subsequence of object position vectors  $\mathbf{s}_{1:t_1,n}$ is initialized with the same values as for the mean vector. Then, we run the DVAE-UMOT algorithm on the first subsequence for $I_0$ iterations. Next, we initialize the mean vector sequence $\mathbf{m}_{t_1+1:t_2,n}$ of the second subsequence with $\mathbf{m}_{t_1 n}$ repeated for $t_2-t_1$ times (and the same for  $\mathbf{s}_{t_1+1:t_2,n}$). And so on for the following subsequences. Finally, the initialized subsequences are concatenated together to form the initialized whole sequence. The pseudo-code of the cascade initialization can be found in Algorithm~\ref{algo2}.
\begin{algorithm}[!t]
\renewcommand{\algorithmicrequire}{\textbf{Input:}}
\renewcommand{\algorithmicensure}{\textbf{Output:}}
\algsetblock[Name]{Initialization}{Stop}{1}{0.5cm}
\caption{Cascade initialization of the position vector sequence}\label{algo2}
\begin{algorithmic}[1]
\Require 
\Statex Detected bounding boxes at the first frame $\mathbf{o}_{1, 1:K_1}$;
\Statex Pre-trained DVAE parameters \{$\theta_{\mathbf{s}}$, $\theta_{\mathbf{z}}$, $\phi_{\mathbf{z}}$\};
\Statex Initialized observation model covariance matrices $\{\boldsymbol{\Phi}^{(0)}_{tk}\}_{t, k=1}^{T, K_t}$;
\Statex Initialized covariance matrices $\{\mathbf{V}^{(0)}_{tn}\}_{t, n=1}^{T, N}$;
\Ensure 
\Statex Initialized mean position vector sequence $\{\mathbf{m}^{(0)}_{tn}\}_{t,n=1}^{T, N}$;
\Statex Initialized sampled position vector sequence $\{\mathbf{s}^{(0)}_{tn}\}_{t,n=1}^{T, N}$;
\State Split the whole observation sequence $\mathbf{o}$ into $J$ sub-sequences indexed by $\{t_0=1, ..., t_1\}$, $\{t_1+1, ..., t_2\}$, ..., $\{t_{J-1}+1, ..., t_J=T\}$;
\For{$j == 1$}
    \For{$k \leftarrow 1$ to $K_1$}
        \State $n \leftarrow k$;
        \For{$t \leftarrow 1$ to $t_1$}
            \State $\mathbf{m}^{(0)}_{tn}, \mathbf{s}^{(0)}_{tn} = \mathbf{o}_{1k}$;
        \EndFor
    \EndFor
\EndFor
\For{$j \leftarrow 2$ to $J$}
    \For{$n \leftarrow 1$ to $N$}
        \For{$t \leftarrow t_{j-1}+1$ to $t_j$}
            \State $\mathbf{m}^{(0)}_{tn}, \mathbf{s}^{(0)}_{tn} = \mathbf{m}^{(I_0)}_{t_{j-1}n}$;
        \EndFor
        \State $\{\mathbf{m}^{(I_0)}_{tn}\}_{t=t_{j-1}+1}^{t_j}$ = \textbf{MOT}($I_0$, $\{\{\theta_{\mathbf{s}}, \theta_{\mathbf{z}}, \phi_{\mathbf{z}}\},$
        \State \multiline{$  \{\boldsymbol{\Phi}^{(0)}_{tk}, \mathbf{m}^{(0)}_{tn},\mathbf{V}^{(0)}_{tn}, \mathbf{s}^{(0)}_{tn}\}_{t=t_{j-1}+1, k=1, n=1}^{t_j, K_t, N}\}$);}
    \EndFor

\EndFor
\State \multiline{ $\mathbf{m}^{(0)}_{1:T,1:N} = \big[\mathbf{m}^{(0)}_{1:t_1,1:N}, ..., \mathbf{m}^{(0)}_{t_{J-1}+1:T,1:N}\big]$;}
\State \multiline{$\mathbf{s}^{(0)}_{1:T, 1:N} = \big[\mathbf{s}^{(0)}_{1:t_1, 1:N}, ..., \mathbf{s}^{(0)}_{t_{J-1}+1:T, 1:N}\big]$;}
\end{algorithmic}
\end{algorithm}

\section{SRNN implementation details}\label{appe:srnn-implementation}
In our MOT set-up, both $\mathbf{s}_t$ and $\mathbf{z}_t$ are of dimension 4. 
The SRNN generative distributions in the right-hand side of \eqref{eq:SRNN-generative-model} are implemented as:
\[
    \mathbf{h}_t = d_h(\mathbf{s}_{t-1}, \mathbf{h}_{t-1}),\label{eq:SRNN-d_h}
\]
\[
    \big[\boldsymbol{\mu}_{\theta_{\mathbf{z}}}, \boldsymbol{v}_{\theta_{\mathbf{z}}} \big] = d_z(\mathbf{h}_t, \mathbf{z}_{t-1}),\label{eq:SRNN-d_z}
\]
\[
    p_{\theta_{\mathbf{z}}}(\mathbf{z}_{t} | \mathbf{s}_{1:t-1}, \mathbf{z}_{t-1}) = \mathcal{N}\big(\mathbf{z}_{t}; \boldsymbol{\mu}_{\theta_{\mathbf{z}}}, \textrm{diag}(\boldsymbol{v}_{\theta_{\mathbf{z}}})\big),\label{eq:SRNN-gen-z}
\]
\[
    \big[\boldsymbol{\mu}_{\theta_{\mathbf{s}}}, \boldsymbol{v}_{\theta_{\mathbf{s}}} \big] = d_s(\mathbf{h}_t, \mathbf{z}_t),\label{eq:SRNN-d_s}
\]
\[
    p_{\theta_{\mathbf{s}}}(\mathbf{s}_{t} | \mathbf{s}_{1:t-1}, \mathbf{z}_t) = \mathcal{N}\big(\mathbf{s}_{t}; \boldsymbol{\mu}_{\theta_{\mathbf{s}}}, \textrm{diag}(\boldsymbol{v}_{\theta_{\mathbf{s}}})\big),\label{eq:SRNN-gen-s}
\]
where the function $d_h$ in (\ref{eq:SRNN-d_h}) is implemented by a forward RNN and $\mathbf{h}_t$ denotes the RNN hidden state vector, the dimension of which is set to 8. In practice, LSTM networks are used. The function $d_s$ in \eqref{eq:SRNN-d_s} is implemented by a dense layer of dimension 16, with the tanh activation function, followed by a linear layer, which outputs are the parameters $\boldsymbol{\mu}_{\theta_{\mathbf{s}}}, \boldsymbol{v}_{\theta_{\mathbf{s}}}$. The function $d_z$ in \eqref{eq:SRNN-d_z} is implemented by two dense layers of dimension 8, 8 respectively, with the tanh activation function, followed by a linear layer, which outputs are the parameters $\boldsymbol{\mu}_{\theta_{\mathbf{z}}}, \boldsymbol{v}_{\theta_{\mathbf{z}}}$. 

\begin{figure*}[!ht]
  \centering
  \includegraphics[width=\linewidth]{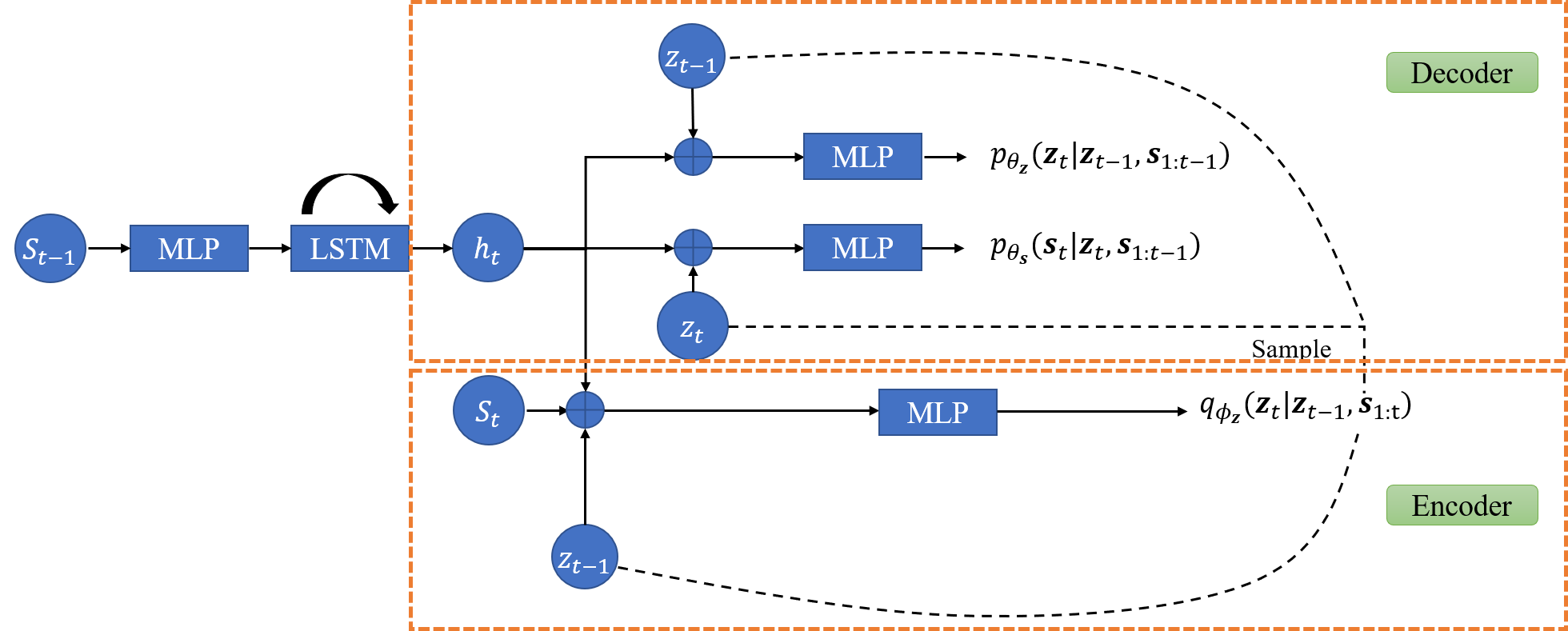}
  \centering
  \caption{Schema of the SRNN model architecture. The ``plus'' symbol represents the concatenation of the input vectors.}
  \label{fig:figure3}
\end{figure*}

The SRNN inference model in the right-hand side of \eqref{eq:SRNN-inference-model-causal} is implemented as:
\newline
\[
    \big[\boldsymbol{\mu}_{\phi_{\mathbf{z}}}, \boldsymbol{v}_{\phi_{\mathbf{z}}}\big] = e_\mathbf{z}(\mathbf{h}_t, \mathbf{s}_t, \mathbf{z}_{t-1}),\label{eq:SRNN-e_z}
\]
\[
    q_{\phi_{\mathbf{z}}}(\mathbf{z}_t|\mathbf{z}_{t-1}, \mathbf{s}_{1:t}) = \mathcal{N}\big(\mathbf{z}_{t}; \boldsymbol{\mu}_{\phi_{\mathbf{z}}}, \textrm{diag}(\boldsymbol{v}_{\phi_{\mathbf{z}}})\big),\label{eq:SRNN-inference-Gaussian-causal}
\]
where the function $e_\mathbf{z}$ in \eqref{eq:SRNN-e_z} is implemented by two dense layers of dimension 16 and 8 respectively, with the tanh activation function, followed by a linear layer, which outputs are the parameters $\boldsymbol{\mu}_{\phi_{\mathbf{z}}}, \boldsymbol{v}_{\phi_{\mathbf{z}}}$. 

The SRNN architecture is schematized in \figurename~\ref{fig:figure3}. It can be noted that the RNN internal state $\mathbf{h}_t$ cumulating the information on $\mathbf{s}_{1:t-1}$ is shared by the encoder and the decoder, see \cite[Chapter 4]{MAL-089} for a discussion on this issue.

\section{Synthetic trajectory dataset generation}\label{appe:synthetic-dataset}
We generate the trajectory of each coordinate of the bounding box, namely $x_t^L$, $x_t^T$, $w_t$ and $h_t$, to from the bounding box sequence. While the trajectory of one coordinate is generated using piece-wise combinations of elementary dynamic functions, which are:  static $a(t)=a_0$, constant velocity $a(t)=a_1t+a_0$, constant acceleration $a(t)=a_2t^2+a_1t+a_0$ and sinusoidal (allowing for circular trajectories) $a(t)=a\sin(\omega t+\phi_0)$. That's to say, we split the whole sequence into several segments, and each segment is dominated by a certain elementary dynamic function. 

The number of segments $s$ is first uniformly sampled in the set $\{1,\ldots,s_{\max}\}$. We then sample $s$ segment lengths that sum up to $T$. This defines the segment boundaries $t_1,\ldots,t_{s-1}$. For each segment, one of the four elementary functions is randomly selected. The function parameters are sampled as follow: $a_1 \sim \mathcal{N}(\mu_{a_1}, \sigma_{a_1}^2)$, $a_2 \sim \mathcal{N}(\mu_{a_2}, \sigma_{a_2}^2)$, $\omega \sim \mathcal{N}(\mu_{\omega}, \sigma_{\omega}^2)$ and $\phi_0 \sim \mathcal{N}(\mu_{\phi_0}, \sigma_{\phi_0}^2)$. The two remaining parameters, $a_0$ and $a$, are set to the values needed to ensure continuous trajectories, thus initialising the trajectories at every segment, except for the first one. The very initial trajectory point is sampled randomly from $ \mathcal{U}(0, 1)$. And the initial width is sampled from a log-normal distribution $w_0 \sim \log \mathcal{N}(\mu_{w_0}, \sigma_{w_0}^2)$. Finally, the ratio between the height and width is supposed to be constant with respect to time. It is sampled from a log-normal distribution $r_{hw} = \frac{h}{w} \sim \log\mathcal{N}(\mu_r, \sigma_r^2)$ and the height is obtained by multiplying the width and the ratio. More implementation details can be found in Algorithm~\ref{algo3}.

\begin{algorithm}[!t]
\renewcommand{\algorithmicrequire}{\textbf{Input:}}
\renewcommand{\algorithmicensure}{\textbf{Output:}}
\algsetblock[Name]{Initialization}{Stop}{1}{0.5cm}
\caption{Synthetic object moving trajectories generation}\label{algo3}
\begin{algorithmic}[1]
\Require 
\Statex Total sequence length $T$;
\Statex Maximum sub-sequence number $s_{max}$;
\Statex Distribution parameters $\mu_{w_0}$, $\sigma_{w_0}$, $\mu_r$, $\sigma_r$, $\mu_{a_1}$, $\sigma_{a_1}$, $\mu_{a_2}$, $\sigma_{a_2}$, $\mu_{\omega}$, $\sigma_{\omega}$, $\mu_{\phi_0}, \sigma_{\phi_0}$;
\Statex Discrete probability distribution of different elementary trajectory function types $p = [p_1, p_2, p_3, p_4]$;
\Ensure 
\Statex Synthetic bounding box position sequence $gen\_seq  = \{(x_t^L, x_t^T, x_t^R, x_t^B)\}_{t=1}^T$;
\Function{GenSeq}{$x_0$, $s$, $t_{split}$, $params\_prob$, $p$}
\State $start = x_0$;
\For{$i \leftarrow 0$ to $s$}
    \State Sample $function\_type$ using $p$;
    \State \multiline{Sample trajectory function parameters $params\_list$ using         $params\_prob$;}
    \State $t_i = t_{split}[i]$;
    \State $x\_sub_i$ = \textbf{GenTraj}($start$, $func\_type$,
    \State $params\_list$);
    \State $start = x\_sub_i[t_i]$;
\EndFor
\State $x = [x\_sub_0, ..., x\_sub_{s-1}]$;
\State \textbf{return} x;
\EndFunction
\State Sample $x_0$, $y_0$ from $\mathcal{U}(0, 1)$;
\State Sample $w_0$ from $\log \mathcal{N}(\mu_{w_0}, \sigma_{w_0})$;
\State Sample $r_{hw}$ from $\mathcal{N}(\mu_r, \sigma_r)$;
\State Randomly sample $s$ in $\{0, ..., s_{max}\}$;
\State Randomly sample $t_{split} = \{t_0, ..., t_{s-1}\}$ in $\{1, ..., T\}$;
\State $x = $ \textbf{GenSeq}({$x_0$, $s$, $t_{split}$, $params\_prob$, $p$});
\State $y = $ \textbf{GenSeq}({$y_0$, $s$, $t_{split}$, $params\_prob$, $p$});
\State $w = $ \textbf{GenSeq}({$w_0$, $s$, $t_{split}$, $params\_prob$, $p$});
\State  $h = w * r_{hw}$;
\State $gen\_seq = [x, y, x+w, y-h]$;
\end{algorithmic}
\end{algorithm}

In our experiments, the total sequence length of the generated trajectories equals to $T=60$ frames. And the maximum number of segments is set to $s_{max}=3$. The parameters of the $a_1$, $a_2$, $\omega$, $\phi_0$, $w_0$, and $r_{hw}$ distributions are determined by estimating the statistical characteristics of publicly published detections of the MOT17 training dataset. More precisely, we estimated the empirical mean and standard deviation of the speed and acceleration for all matched detection sequences (i.e., the first and second order differentiation of the position sequences). 

\end{document}